\documentclass[runningheads]{llncs}
\usepackage{graphicx}
\usepackage{subfig}
\usepackage{xspace}
\usepackage{tikz}
\usepackage{comment} 
\usepackage{amsmath,amssymb} %
\usepackage{color}
\newcommand{\ignore}[1]{}

\usepackage{multirow}
\usepackage{array}
\usepackage{booktabs}
\usepackage{wrapfig}

\usepackage{soul}
\usepackage{url}
\usepackage[symbol]{footmisc}

\usepackage[colorlinks=true,citecolor=green]{hyperref}

\usepackage{colortbl}

\usepackage[toc,page]{appendix}

\usepackage{amsmath,amsfonts,bm}

\def\eqref#1{equation~\ref{#1}}

\def\1{\bm{1}}

\newcommand{\test}{\mathcal{D_{\mathrm{test}}}}

\DeclareMathAlphabet{\mathsfit}{\encodingdefault}{\sfdefault}{m}{sl}
\SetMathAlphabet{\mathsfit}{bold}{\encodingdefault}{\sfdefault}{bx}{n}

\newcommand{\prithvi}{\textcolor{black}}

\newcommand{\model}{DMG\xspace}

\makeatletter
\DeclareRobustCommand\onedot{\futurelet\@let@token\@onedot}
\def\@onedot{\ifx\@let@token.\else.\null\fi\xspace}

\makeatother

\newcommand{\eat}[1]{}
\newcommand{\replace}[2]{} %

\begin{document}
\pagestyle{headings}
\mainmatter
\def\ECCVSubNumber{790}  %

\title{Learning to Balance Specificity and Invariance \\
for In and Out of Domain Generalization} %

\titlerunning{Domain Specific Masks for Generalization}
\author{Prithvijit Chattopadhyay\inst{1} \and Yogesh Balaji\inst{2} \and Judy Hoffman\inst{1}}
\authorrunning{Chattopadhyay et al.}
\institute{Georgia Institute of Technology \and University of Maryland \\
\email{\{prithvijit3,judy\}@gatech.edu, yogesh@cs.umd.com}}
\maketitle

\begin{abstract}
We introduce \textbf{D}omain-specific \textbf{M}asks for \textbf{G}eneralization, a model for improving both in-domain and out-of-domain generalization performance. For domain generalization, the goal is to learn from a set of source domains to produce a single model that will best generalize to an unseen target domain. As such, many prior approaches focus on learning representations which persist across all source domains with the assumption that these domain agnostic representations will generalize well. However, often individual domains contain characteristics which are unique and when leveraged can significantly aid in-domain recognition performance. To produce a model which best generalizes to both seen and unseen domains, we propose learning domain specific masks. The masks are encouraged to learn a balance of domain-invariant and domain-specific features, thus enabling a model which can benefit from the predictive power of specialized features while retaining the universal applicability of domain-invariant features. We demonstrate competitive performance compared to naive baselines and state-of-the-art methods on both PACS and DomainNet.\footnote{Our code is available at \href{https://github.com/prithv1/DMG}{https://github.com/prithv1/DMG}}
\end{abstract}

\section{Introduction}

The success of deep learning has propelled computer vision systems from purely academic endeavours to key components of real-world products.
This deployment into unconstrained domains has forced researchers to focus attention beyond a closed-world supervised learning paradigm, where learned models are only evaluated on held-out in-domain test data, and instead produce models capable of generalizing to diverse test time data distributions. 

This problem has been formally studied and progress measured in the \textit{domain generalization} literature~\cite{muandet2013domain,ghifary2015domain}.  
Most prior work in domain generalization focuses on learning a model which generalizes to unseen domains by either directly optimizing for domain invariance~\cite{muandet2013domain} or designing regularizers that induce such a bias~\cite{balaji2018metareg}, the idea being that features which are present across multiple training distributions are more likely to persist in the novel distributions.
However, in practice, as the number of training time data sources increases it becomes ever more likely that at least some of the data encountered at test time will be very similar to one or more source domains. In such a situation, ignoring features specific to only a domain or two may artificially limit the efficacy of the final model. However, leveraging a balance between ``\textit{invariance}'' -- features
that are shared across domains -- and ``\textit{specificity}'' -- features which are specific to
individual domains -- might actually aid the model in
making a better prediction.

\begin{figure}[t]
\centering
\includegraphics[width=0.7\textwidth]{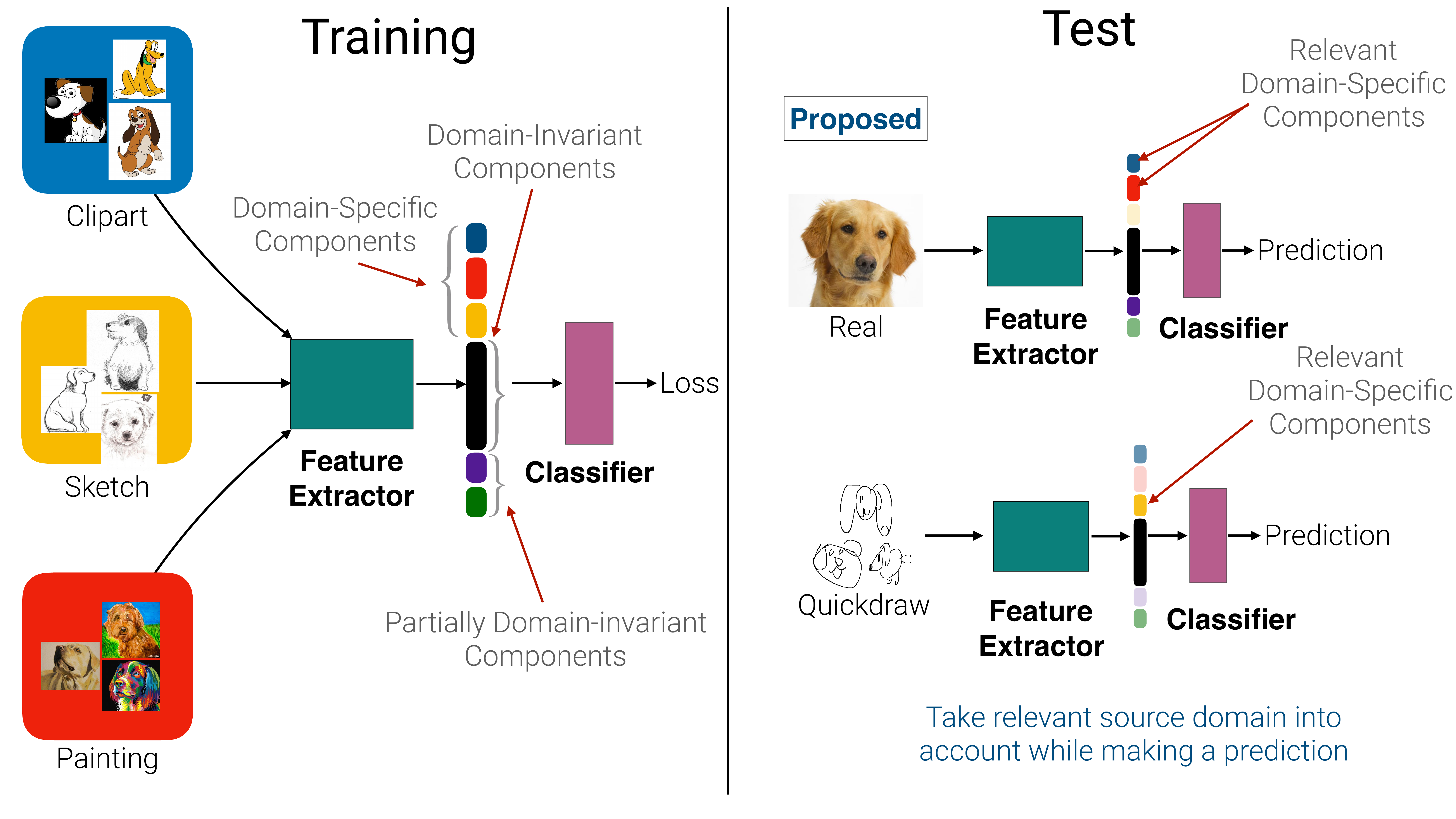}
\caption{\textbf{Balancing specificity and invariance.} At training time, we optimize for a combination of domain-specific (shown in blue, yellow, red) and domain invariant (shown in black) learned representations. Partially invariant representations are indicated as color combinations (i.e. blue + yellow = green).     
At test-time, these learned representations that capture a balance
of domain-specificity and invariance
 allow the classifier to make a better prediction for given test-instance by leveraging domain-specific features from the most similar source domains.  }
\label{fig:teaser_fig}
\end{figure}

It is important to note that the similarity of data
encountered at test-time to a source domain can be understood clearly
only in the context of the other available source domains.
Consider the example in Fig.~\ref{fig:teaser_fig}, where a classifier trained
on \textit{clipart}, \textit{sketch} and \textit{painting} encounters an 
instance from a novel domain \textit{quickdraw} at test-time. Due to the severe domain-shift involved, leveraging the
relative similarity of the test-instance to samples from \textit{sketch} 
might result in a better prediction compared to a setting where the model
relies solely on invariant characteristics across domains.
However, manually crafting such a balance
or creating an explicit separation between domain-specificity
and invariance~\cite{khosla2012undoing} is not scalable as the
number and diversity of the source distributions available during training increases.

In this paper, we propose 
\model :
\textbf{D}omain-specific \textbf{M}asks for \textbf{G}eneralization,
an algorithm for automatically learning to balance between domain-invariant and domain-specific features producing a single model capable of simultaneously achieving strong performance across multiple distinct domains. %
At a high-level, we cast this problem of \textit{balanced} feature selection as one of learning
distribution-specific binary masks over features of a shared deep
convolutional network (CNN).
Specifically, for a given layer in the CNN,
we associate domain-specific mask parameters
for each neuron which decide whether to turn that neuron
\textit{on} or \textit{off} during a forward pass.
We learn these masks end-to-end via backpropagation
along with the network parameters.
To promote discriminative features and strong end-task performance, we
simultaneously minimize the standard classification error and, 
to encourage domain-specificity in the selected features, we penalize for
overlap amongst masks from different source domains. 
Importantly, our approach uses straightforward optimization across all pooled source data without any need for multi-stage training or meta-learning. 
At test-time we average the predictions obtained by applying all the individual source domain masks thus making a prediction that is informed by both characteristics which are shared across
the source domains and are specific to individual domains. Based on our experiments, we find that not only does
our modeling choice result in at par or improved performance compared to other 
complex
alternatives that explicitly
model domain-shift during training, but also allows us to explicitly characterize activations specific to individual
source domains. Compared to prior work, we find that our approach is much more scalable and is faster to train as training time is essentially
equivalent
to the same as training a vanilla aggregate baseline which pools data from
multiple source domains and trains a single deep network.

Additionally, we note that efforts towards domain generalization in the computer vision literature have focused primarily on measuring novel domain performance at test time. 
Since it is likely that in a realistic scenario the model might also encounter data from the 
source distributions at test-time, it is equally important to \textit{retain}
strong performance on the source distributions in addition to improved generalization to novel domains. Thus, given that
measuring continued holistic progress 
in
domain generalization requires benchmarking
proposed solutions in terms of both in and out-of-domain generalization performance, we also report 
in-domain generalization performance on the large DomainNet~\cite{peng2019moment} benchmark proposed for
domain adaptation. Concretely, we make the following contributions.
\begin{itemize}
    \item We introduce an approach, DMG: \textbf{D}omain-specific \textbf{M}asks for \textbf{G}eneralization, that learns models capable of balancing specificity and invariance over multiple distinct domains.
    We demonstrate that despite our relatively simple approach, \model achieves competitive out-of-domain performance on the commonly used PACS~\cite{Li2017dg} benchmark and on the challenging DomainNet~\cite{peng2019moment} dataset. In addition, we demonstrate that our model can be used as a drop-in replacement for an aggregate model when evaluated on in-domain test samples, or can be trivially converted into a high performing domain-specific model given a known test time domain label.
    
    \item  We verify that our model does indeed lead to the emergence of domain specificity and show that our test time performance is stable across a variety of allowed domain overlap settings. Though not the focus of this paper, this domain specificity may be a helpful tool towards model interpretability. 
\end{itemize}

\section{Related Work}

\par \noindent
\textbf{Domain Adaptation.} Significant progress has been made in the problem of unsupervised domain adaptation where
given access to a labeled source and an unlabeled target dataset,
the task is to improve performance on the target domain. One popular line of approaches include learning a domain invariant representation by minimizing
the distributional shift between source and target feature distributions using an adversarial loss~\cite{ganin2016domain,tzeng2017adversarial}, or MMD-based loss~\cite{long2015DAN,long2016RTN,Long2017JAN}.  While these approaches perform alignment in the feature space, pixel-level alignment 
is performed 
using cross-domain generative models such as GANs in \cite{Bousmalis_2017_CVPR}. A combination of feature-level and pixel-level aligment is explored in \cite{hoffman2018cycada,sankaranarayanan2018GTA}. In addition, several regularization strategies have also been proven to be effective for domain adaptation such as dropout regularization~\cite{saito2018adversarial}, classifier discrepancy~\cite{saito2017maximum}, self-ensembling~\cite{french2018selfensembling}, etc. Most existing domain adaptation methods
consider the setting where the source and the target datasets contain one domain each. In multi-source domain adaptation, the source dataset consists of a mixture of multiple domains
where domain alignment is performed using an adversarial interplay involving a $k$-way domain discriminator in \cite{xu2018deep}, and multi-domain moment matching in \cite{peng2019moment}.

\par \noindent
\textbf{Domain Generalization.} Similar to the multi-source domain adaptation problem, domain generalization considers multiple domains in the input data distribution. However, no access to the target distribution (including the unlabeled target) is assumed during training. This makes domain generalization a much harder problem than multi-source adaptation. One common approach to the problem involves decomposing a model into domain-specific and domain-invariant components, and using the domain-invariant component to make predictions at test time~\cite{MTAE,Undo_bias}. Recently, the use of meta-learning for domain generalization has gained much attention. \cite{li2018learning} extends the MAML framework of \cite{finn2017model} for domain generalization by learning parameters that adapt quickly to target domains. In \cite{balaji2018metareg}, a regularization function is estimated using meta-learning, which when used with multi-domain training results in a robust minima with improved domain generalization. Use of data augmentation techniques for domain generalization is explored in \cite{volpi2018generalizing}. Recently, a novel variant of empirical risk minimization framework, called Invariant Risk Minimization (IRM)
has been proposed in \cite{arjovsky2019invariant,ahuja2020invariant} to make machine learning models invariant to spurious correlations in data when training across multiple sources. 

\par \noindent
\textbf{Disentangled Representations.} The goal of learning disentangled representations is to be able to disentangle learned features into multiple factors of variations, each factor representing a semantically meaningful concept. The problem has primarily been studied in the unsupervised setting. Typical approaches involve training a generative model such as a GAN or VAE while imposing constraints in the latent space using KL-divergence~\cite{pmlr-v80-kim18b,burgess2018understanding} or mutual information~\cite{NIPS2016_6399}. In the context of domain adaptation, disentangling features into domain-specific and domain-independent factors have been proposed in \cite{bousmalis2016domain,Peng2019DomainAL}. The domain-independent factors are then used to obtain predictions in the target domain. 
Our approach performs a similar implicit disentanglement, 
where domain-specific and domain-invariant factors are mined using a masking operation.

\par \noindent
\textbf{Dropout, Pruning, Sparsification and Attention.} Our approach to learn domain-specific masks is similar to the techniques adopted in the network pruning and sparsification literature. 
Relevant to our work are approaches
that directly learn a pruning strategy during training~\cite{savarese2019winning,srivastava2014dropout,venkatesh2020calibrate}.~\cite{savarese2019winning} involves learning masks over parameters under a sparsity constraint to discover small sub-networks. 
In addition to model compression, pruning strategies have also been used in multi-task and continual learning. In~\cite{serra2018overcoming}, catastrophic forgetting is prevented while learning
tasks (and subsequently attending over them) in a sequential manner. 
In \cite{mallya2018piggyback}, a binary mask corresponding to individual tasks are learnt for a fixed backbone network. The resulting task-specific network is obtained by applying the learnt masks on the backbone network. In \cite{mallya2018packnet}, weights of a network are iteratively pruned to free up packets of neurons. The free neurons are in-turn updated to learn new tasks without forgetting. A similar approach is proposed in \cite{berriel2019budget} for multi-domain learning where domain-specific networks are constructed by masking convolution filters under a budget on new parameters being introduced for each domain.
Similarly, several approaches building on top of Dropout~\cite{srivastava2014dropout}
have also been proposed 
for domain adaptation. In \cite{saito2018adversarial}, a pair of sub-networks are sampled from dropout that give maximal classifier discrepancy. Feature network is trained to minimize this discrepancy, thus making it insensitive to perturbations in classifier weights. An efficient implementation of this idea using adversarial dropout is proposed in \cite{lee2019drop}.
In~\cite{zunino2020explainable}, saliency supervision is used to develop
explainable models for domain generalization.
While \model is akin to
attention being used as learned masks for subset selection~\cite{mallya2018packnet,mallya2018piggyback}, our focus is on
implicitly learning to disentangle domain-specific and invariant feature
components for multi-source domain generalization.

\section{Approach}
Our motivation to ensure a balance between specificity and invariance is to aid prediction
in situations where an instance at test-time might benefit from some of the domain-specific components
captured by the domain-specific masks. In what follows, we first describe the problem setup, ground associated notations and then describe our proposed approach, DMG.

\subsection{Problem Setup}
Domain generalization involves 
training a model on data, denoted as $\mathcal{X}$, sampled from $p$ source distributions that generalizes
well to $q$ unknown target distributions which lack training data. 
Without loss of generality we focus on the classification case, where the goal is to learn a model which maps inputs to the desired output label, $M: \mathcal{X}\rightarrow \mathcal{Y}$. 
 Let $\{\mathcal{D}_i\}_{i=1}^{p+q}$
denote the $p+q$ distributions with same support $\mathcal{X}\times\mathcal{Y}$.
Let $D_i=\{(x_j^{(i)}, y_j^{(i)})\}_{i=1}^{|D_i|}$ refer to the dataset
sampled from the $i^{th}$ distribution, i.e., $D_i \sim \mathcal{D}_i$. 
We operate in the setting where all the distributions share the same
label space and distributional variations exist only in the input data (space $\mathcal{X}$).
We are interested in learning a parametric model $M_{\Theta}:\mathcal{X}\rightarrow\mathcal{Y}$, that we can decompose into a
feature extractor ($F_{\psi}$) and a task-network ($T_{\theta}$) i.e., 
$M_{\Theta}(x) = (T_{\theta} \circ F_{\psi}) (x)$, where $\Theta, \psi, \theta$
denote the parameters of the complete, feature and the task networks
respectively. For the remaining subsections, we refer to the set of
source domains as $D_S$ and index individual source domains by $d$.
We learn domain specific masks only on the
neurons present in the task network.

\begin{figure}[t]
    \begin{center}
    \centering
  \includegraphics[width=0.8\textwidth]{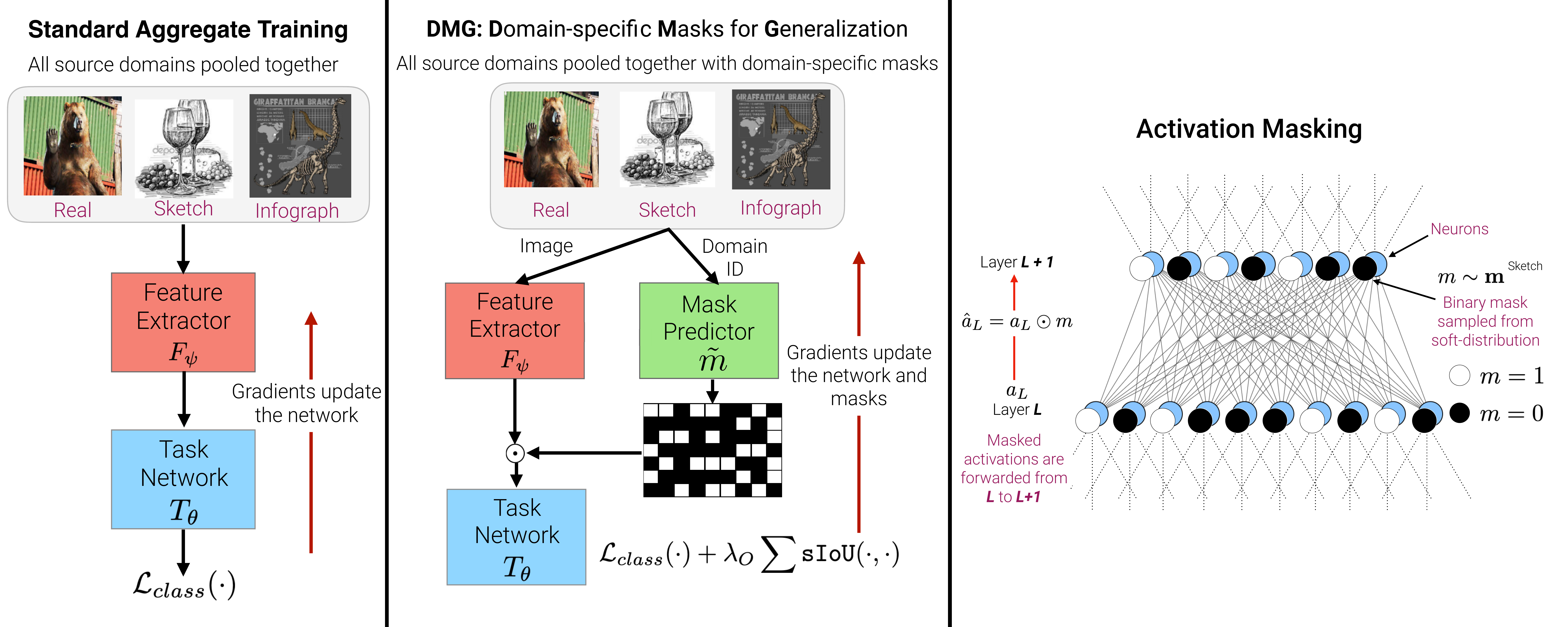}
	\caption{\textbf{Illustration of our approach (\model)}: We introduce domain-specific activation masks for learning a balance between domain-specific and domain-agnostic features. [Left] Our training pipeline
    involves incorporating domain-specific masks in the vanilla aggregate training process. [Middle] For an image
    belonging to \textit{sketch}, we sample a binary mask from the corresponding mask parameters, which is then applied to the neurons of the task-network.
    \ignore{to be applied to the neurons of the task-network from the corresponding mask parameters. }
    [Right] Post feature extraction, an elementwise product of the obtained binary masks is performed
    with the neurons of the task network layer ($L$) to obtain the \textit{effective} activations being 
    passed on to the next layer ($L+1$). The mask and network parameters are learned end-to-end
    based on the standard cross-entropy coupled with the \texttt{sIoU} loss penalizing mask overlap
    among the source domains.}
	\label{fig:train_pipeline}
\end{center}
\end{figure}

\subsection{Activation or Feature Selection via Domain-Specific Masks}
\label{sec:feature_selection}
Our goal is to learn representations which capture a balance of
domain specific components (useful for predictive performance on a specific
domain) and domain invariant components (useful in general for the discriminative task).
Capturing information contained in \ignore{the} multiple source distributions in such
a manner allows us to make better predictions by automatically relying \ignore{on} more on \ignore{the specific}
 characteristics of a specific source domain in situations where an
instance observed at test-time is relatively similar to one of the sources\ignore{domains}.
We cast this problem of disentangling domain-specific and domain-invariant feature components as that of learning binary masks on the neurons of the task network specific to individual source domains.
\ignore{We cast this problem of capturing domain-specific and domain-invariant as one
of binary masks over neurons of the task network specific to individual source domains.}
More specifically, for each of the $p$ source distributions, we initialize masks $\mathbf{m}^d$
over neurons (or activations) of the task-network $T_{\theta}$. Our masks can
be viewed as layer-wise gates 
which decide which 
neurons
to turn
\textit{on} or \textit{off} during a forward pass through the network.

Given $k$
neurons at some layer $L$ of $T_{\theta}$, we introduce parameters $\Tilde{\mathbf{m}}^d \in \mathbb{R}^k$ for each of the 
source distributions $d \in D_S$.
\ignore{
During training, for some instance $x_i^d$ from domain $d$,
we treat the binary masks as stochastic samples from the bernoulli distribution $\mathbf{m}^d = \sigma(\Tilde{\mathbf{m}}^d)$\footnote{$\sigma(x) = \frac{1}{1 + \exp(-x)}$} as 
 $m^d_i \sim \mathbf{m}^d$ such that $m^d_i \in \{0,1\}^k$.}
During training, for instances $x_i^d$ from domain $d$, we first form mask probabilities $\mathbf{m}^d$ via a sigmoid operation as $\mathbf{m}^d = \sigma(\Tilde{\mathbf{m}}^d)$.
Then, the binary masks $m_{i}^{d}$ are sampled from a bernoulli distribution given by the mask probabilities. i.e., $m^d_i \sim \mathbf{m}^d$, with $m^d_i \in \{0,1\}^k$.
 Upon sampling
masks for individual neurons, the effective activations which are passed on to the next
layer $L+1$ are $\hat{a}_L = a_{L} \odot m^d_i$, i.e., an elementwise product of the
obtained 
activations and the sampled binary masks (see Fig.~\ref{fig:train_pipeline}, right).
During training, we sample such binary masks corresponding to the 
source domain of the input instance,
thereby making feedforward predictions by only using domain-specific masks. Under this setup, the prediction made by the
entire network $M_{\Theta}$ for an instance $x_i^d \in d$ can be expressed as
$\hat{y}_i = M_{\Theta}(x_i^d; m^d_i)$ where $m^d_i$ denotes the sampled mask (for domain $d$) being applied to all neurons
in the task-network $T_{\theta}$. Note that, akin to dropout~\cite{srivastava2014dropout}, these domain-specific masks identify 
\textit{domain-specific} sub-networks -- for an instance $x_i^d$, the sampled binary
mask $m^d_i$ identifies a specific ``thinner'' subnetwork.

We learn the mask-parameters in addition to the parameters of the network during
training. However, note that the mask-parameters $\Tilde{\mathbf{m}}^d$ cannot
be updated directly using back-propagation as the sampled binary mask is discrete.
We approximate gradients through sampled discrete masks using the straight-through estimator~\cite{bengio2013estimating}, i.e.,
we use a discretized $m^d_i$ during a forward pass but use the continuous version $\mathbf{m}^d$ during the backward pass by approximating 
$\nabla_{m^d_i} \mathcal{L} \approx \nabla_{\mathbf{m}^d} \mathcal{L}$.
Even though the hard sampling step is non-differentiable, gradients with respect to 
$m^d_i$ serve as a noisy estimator of 
$\nabla_{\mathbf{m}^d} \mathcal{L}$.

\par \noindent
\textbf{Incentivizing Domain-Specificity.} To ensure the masks capture neurons that are specific to individual source domains, we need to encourage 
specificity
in the masks while maximizing predictive performance on the
source set of distributions. To incentivize domain-specificity, we introduce an
additional \textit{soft-}overlap loss that ensures masks
associated with each of the source distributions overlap minimally. To quantify overlap 
we compute the 
Jaccard Similarity Coefficient~\cite{article} (also known as IoU score) among pairs of source domain masks.
However, as IoU is non-differentiable it is not possible to directly optimize for the same
using gradient descent.
Therefore, 
inspired by prior work~\cite{rahman2016optimizing},
we minimize the following \textit{soft-}overlap
loss for every pair of source domain masks $\{\mathbf{m}^{d_i}, \mathbf{m}^{d_j}\}$ at a layer $L$ as,
\begin{equation}
    \mathtt{sIoU} (\mathbf{m}^{d_i}, \mathbf{m}^{d_j}) = \frac{\mathbf{m}^{d_i} \cdot \mathbf{m}^{d_j}}{\sum_k (\mathbf{m}^{d_i} + \mathbf{m}^{d_j} - \mathbf{m}^{d_i} \odot \mathbf{m}^{d_j})}
\end{equation}
where $\mathbf{m}^{d_i} \cdot \mathbf{m}^{d_j}$ approximates the intersection 
for the pair
of source domain masks as the inner product of the mask distributions, $\odot$ denotes the
elementwise product and $k$ denotes the number of neurons in layer $L$. During
training \texttt{sIoU} ensures predictions for instances from different source
domains are made using different sub-networks (as identified by the domain-specific
binary masks).

To summarize, for a set of source domains $D_S$ the overall objective
we optimize during training ensures -- (1) good predictive performance on the discriminative task at hand and (2) minimal
overlap among source-domain masks, 
\begin{align}
    \mathcal{L}(\theta, \psi, \Tilde{\mathbf{m}}^{d_1}, .., \Tilde{\mathbf{m}}^{d_{|D_S|}}) = 
    \sum_{d \in D_S}\sum_{x^d_i \in d}\mathcal{L}_{\mathtt{class}} (\theta, \psi, m^d_i) \nonumber \\
    + \lambda_O \sum_{L \in T_{\theta}} \sum_{(d_i,d_j) \in D_S} \mathtt{sIoU}(\mathbf{m}^{d_i}, \mathbf{m}^{d_j})
\label{eq:overall_obj}
\end{align}
where $m^d_i \sim \mathbf{m}^d_i$ for every instance $x^d_i$ of the source domain $d$ and $\mathcal{L}_{\mathtt{class}}(\cdot)$ denotes the standard cross entropy loss. Fig.~\ref{fig:train_pipeline}
summarizes our training pipeline in context of a standard aggregation method where a CNN 
is trained jointly on data pooled from all the source domains.

\par \noindent
\textbf{Prediction at Test-time.} 
To obtain a prediction at test-time, we follow
a soft-scaling scheme similar to Dropout~\cite{srivastava2014dropout}. 
Recall that sampling
from domain-specific \textit{soft-}masks essentially amounts to sampling a ``thinned'' sub-network
from the orginal task-network. 
However,
since it is intractable to obtain predictions from all such possible (exponential) domain-specific sub-networks, we follow a simple averaging scheme that ensures that the \textit{expected} output
    under the distribution induced by the masks is the same as the actual output at test-time.
Specifically, we scale every neuron by the associated domain-specific \textit{soft-}mask
    $\mathbf{m}^d$ instead of turning neurons \textit{on} or \textit{off} based
    on a discrete mask $m\sim\mathbf{m}^d$ and average the predictions 
    obtained by applying $\mathbf{m}^d$ for all the source domains to the task network.\footnote{We experimented with learning a domain-classifier
on source domains to use the predicted probabilities as weights for
test-time averaging. We observed insignificant difference in
out-of-domain performance but significantly worse in-domain performance, though we
believe this may be dataset-specific.}

\section{Experiments}
\subsection{Experimental Settings.} 
\par \noindent
\textbf{Datasets and Metrics.} We conduct domain generalization (DG) experiments
on the following datasets:
\label{para:metrics}

\textbf{PACS}~\cite{Li2017dg} -- PACS is a recently proposed benchmark for domain generalization
which
consists of 
only 9991 images of 7 classes, distributed across 4 domains - \textit{photo}, \textit{art-painting}, \textit{cartoon} and \textit{sketch}. Following standard practice, we conduct 
4 sets of experiments -- treating one domain as the unseen target and the rest as
the source set of domains. The authors of~\cite{Li2017dg} provide specified \texttt{train} and \texttt{val} splits for each domain to ensure fair comparison and treat the 
entirety of \texttt{train} + \texttt{val} as the \texttt{test}-split of the target domain. We use the same splits for our experiments. As such, the proposed splits
do not include an in-domain \texttt{test}-split, thereby limiting us from
computing in-domain performance in addition to measuring out-of-domain generalization.
    
\begin{table}[t]
\renewcommand*{\arraystretch}{1.18}
\setlength{\tabcolsep}{6pt}
\begin{center}
\caption{\textbf{Out of Domain 
Accuracy (\%) on DomainNet ($\lambda_O$ = 0.1)} 
\\ \footnotesize $^\mp$We were unable to optimize the MetaReg~\cite{balaji2018metareg} objective with Adam~\cite{kingma2014adam} as the optimizer and therefore, we also include comparisons with Aggregate and MetaReg trained with SGD.}
\resizebox{0.8\columnwidth}{!}{
\begin{tabular}{l l l  c  c c c c c c | c }
\toprule
& & \textbf{Method} & & C & I & P & Q & R & S & Overall \\
\midrule
\multirow{5}{*}{\rotatebox{90}{\centering AlexNet }} & 
 & Aggregate && 47.17 & 10.15 & 31.82 & 11.75 & 44.35 & 26.33& 28.60\\
& & Aggregate-SGD$^\mp$ && 42.30 & 12.42 & 31.45 & 9.52 & 42.76 & 29.34& 27.97\\
& & Multi-Headed  && 45.96 & 10.56 & 31.07 & 12.05 & 43.56 & 25.93 & 28.19 \\
& & MetaReg~\cite{balaji2018metareg}$^\mp$ && 42.86 & \textbf{12.68} & 32.47 & 9.37 & 43.43 & 29.87 & 28.45\\
\cline{3-11}
& & \model (Ours)  && \textbf{50.06} & 12.23 & \textbf{34.44} & \textbf{13.07} & \textbf{46.98} & \textbf{30.13} & \textbf{31.15} \\
\midrule
\multirow{5}{*}{\rotatebox{90}{\centering ResNet-18 }} & 
& Aggregate  && 57.15 & 17.69 & 43.21 & 13.87 & 54.91 & 39.41 & 37.71\\
& & Aggregate-SGD$^\mp$  && 56.56 & 18.44 & \textbf{45.30} & 12.47 & 57.90 & 38.83 & 38.25\\
& & Multi-Headed && 55.46 & 17.51 & 40.85 & 11.19 & 52.92 & 38.65 & 36.10\\
& & MetaReg~\cite{balaji2018metareg}$^\mp$ && 53.68 & \textbf{21.06} & 45.29 & 10.63 & \textbf{58.47} & \textbf{42.31} & 38.57\\
\cline{3-11}
& & \model (Ours)  && \textbf{60.07} & 18.76 & 44.53 & \textbf{14.16} & 54.72 & 41.73 & \textbf{39.00} \\
\midrule
\multirow{5}{*}{\rotatebox{90}{\centering ResNet-50 }} & 
& Aggregate  && 62.18 & 19.94 & 45.47 & 13.81 & 57.45 & 44.36 & 40.54\\
& & Aggregate-SGD$^\mp$  && 64.04 & 23.63 & \textbf{51.04} & 13.11 & 64.45 & 47.75 & \textbf{44.00}\\
& & Multi-Headed && 61.74 & 21.25 & 46.80 & 13.89 & 58.47 & 45.43 & 41.27\\
& & MetaReg~\cite{balaji2018metareg}$^\mp$ && 59.77 & \textbf{25.58} & 50.19 & 11.52 & \textbf{64.56} & \textbf{50.09} & 43.62\\
\cline{3-11}
& & \model (Ours)  && \textbf{65.24} & 22.15 & 50.03 & \textbf{15.68} & 59.63 & 49.02 & 43.63\\
\bottomrule
\end{tabular}}
\label{tab:keyodres_dmnt}
\end{center}
\end{table}

\textbf{DomainNet}~\cite{peng2019moment} -- 
DomainNet is a recently proposed large-scale dataset
 for domain adaptation
 which 
 consists
 of $\sim$0.6 million images of 345 classes distributed across 6 domains -- \textit{real}, \textit{clipart}, \textit{sketch}, \textit{painting}, \textit{quickdraw} and \textit{infograph}. 
 DomainNet surpasses all prior 
 datasets for domain adaptation significantly in terms of size and diversity.
 The authors of~\cite{peng2019moment} recently released annotated \texttt{train} and
 \texttt{test} splits for all the 6 domains. We 
 divide
 the \texttt{train} split from~\cite{peng2019moment}
 randomly in a 90-10$\%$ proportion to obtain \texttt{train} and \texttt{val}
 splits for our experiments. Similar to PACS, we conduct 6 sets of leave-one-out experiments.
We report out-of-domain performance as the accuracy on the
\texttt{test} split of the unseen domain. For in-domain
performance, we report accuracy averaged over all the source domain \texttt{test} splits.

\par \noindent
\textbf{Models.} We experiment with 
ImageNet~\cite{deng2009imagenet} pretrained
AlexNet~\cite{krizhevsky2012imagenet}, ResNet-18~\cite{he2016deep} and ResNet-50~\cite{he2016deep} backbone architectures.
For AlexNet, we apply
domain-specific masks on the input activations of the last three fully-connected 
layers -- our task network $T_{\theta}$ -- and turn dropout~\cite{srivastava2014dropout} \textit{off} while learning the
domain-specific masks. For ResNet-18 and 50, we apply domain specific masks on the
input activations of the last residual block and the first fully connected layer.\footnote{Specifically, for ResNet, the
domain-specific masks are trained to \textit{drop} or \textit{keep} specific channels in the input activations
as opposed to every spatial feature in every channel 
in order to
reduce 
complexity in terms of 
the number of mask parameters to be learnt.}

\par \noindent
\textbf{Baselines and Points of Comparison.} We 
compare \model with two simple baselines (treating dropout~\cite{srivastava2014dropout} as usual if present in the backbone CNN) -- (1) Aggregate - the CNN backbone trained jointly on data
accumulated from all the source domains and (2) Multi-Headed - the
CNN backbone with different classifier heads corresponding to each of the source domains (at test-time we average predictions from all the classifier heads). Note, this baseline has more parameters than our model due to the repeated classification heads. In addition to the
above baselines, we also compare with the recently proposed domain generalization approaches
(cited in Tables~\ref{tab:keyodres_dmnt},\ref{tab:keyodres_pacs} and~\ref{tab:keyidres_dmnt}). 
Please refer to the appendix for implementation details.

\begin{table}[t] 
\renewcommand*{\arraystretch}{1.18}
\setlength{\tabcolsep}{6pt}
\begin{center}
\caption{\textbf{Out of Domain 
Accuracy (\%) on PACS ($\lambda_O$ = 0.1)} 
\\ \footnotesize $^*$We include the aggregate baseline both as reported in~\cite{li2019episodic} as well as our own implementation (indicated as Aggregate$^*$).}
\resizebox{0.6\columnwidth}{!}{
\begin{tabular}{l l l  c  c c c c | c }
\toprule
& & \textbf{Method} & & A & C & P & S & Overall \\
\midrule
\multirow{11}{*}{\rotatebox{90}{\centering AlexNet }} & 
& Aggregate~\cite{li2019episodic} && 63.40 & 66.10 & 88.50 & 56.60 & 68.70 \\
& & Aggregate* && 56.20 & 70.69 & 86.29 & 60.32 & 68.38 \\
& & Multi-Headed && 61.67 & 67.88 & 82.93 & 59.38 & 67.97 \\
& & DSN~\cite{bousmalis2016domain} && 61.10 & 66.50 & 83.30 & 58.60 & 67.40 \\
& & Fusion~\cite{mancini2018best} && 64.10 & 66.80 & 90.20 & 60.10 & 70.30\\
& & MLDG~\cite{li2018learning} && 66.20 & 66.90 & 88.00 & 59.00 & 70.00 \\
& & MetaReg~\cite{balaji2018metareg} && 63.50 &	69.50 &	87.40 & 59.10 & 69.90 \\
& & CrossGrad~\cite{volpi2018generalizing} && 61.00 & 67.20 & 87.60& 55.90 &67.90 \\
& & Epi-FCR~\cite{li2019episodic} && 64.70 & 72.30 & 86.10& 65.00 &72.00 \\
& & MASF~\cite{dou2019domain} && \textbf{70.35} & \textbf{72.46} & \textbf{90.68}& 67.33 &\textbf{75.21} \\

\cline{3-9}
& & \model (Ours)  && 64.65 & 69.88 & 87.31 & \textbf{71.42} & 73.32\\
\midrule
\multirow{9}{*}{\rotatebox{90}{\centering ResNet-18 }} & 
& Aggregate~\cite{li2019episodic} && 77.60 & 73.90 & 94.40 & 74.30 & 79.10 \\
& & Aggregate* && 72.61 & 78.46 & 93.17 & 65.20 & 77.36 \\
& & Multi-Headed  && 78.76 & 72.10 & 94.31 & 71.77 & 79.24 \\
& & MLDG~\cite{li2018learning} && 79.50 & 77.30 & 94.30 & 71.50 & 80.70 \\
& & MetaReg~\cite{balaji2018metareg} && 79.50 &	75.40 &	94.30 & 72.20 & 80.40 \\
& & CrossGrad~\cite{volpi2018generalizing} && 78.70 & 73.30 & 94.00 & 65.10 & 77.80 \\
& & Epi-FCR~\cite{li2019episodic} && \textbf{82.10} & 77.00 & 93.90 & 73.00 & \textbf{81.50} \\
& & MASF~\cite{dou2019domain} && 80.29 & 77.17 & \textbf{94.99} & 71.68 & 81.03\\
\cline{3-9}
& & \model (Ours)  && 76.90 & \textbf{80.38} & 93.35 & \textbf{75.21} & 81.46\\
\midrule
\multirow{4}{*}{\rotatebox{90}{\centering ResNet-50 }} & 
& Aggregate* && 75.49 & \textbf{80.67} & 93.05 & 64.29 & 78.38 \\
& & Multi-Headed  && 75.15 & 76.37 & \textbf{95.27} & 75.26 & 80.51 \\
& & MASF~\cite{dou2019domain} && \textbf{82.89} & 80.49 & 95.01 & 72.29 & 82.67\\
\cline{3-9}
& & \model (Ours)  && 82.57 & 78.11 & 94.49 & \textbf{78.32} & \textbf{83.37}\\
\bottomrule
\end{tabular}}
\label{tab:keyodres_pacs}
\end{center}
\end{table}

\begin{table}[t] 
\renewcommand*{\arraystretch}{1.18}
\setlength{\tabcolsep}{6pt}
\begin{center}
\caption{\textbf{In Domain 
Accuracy (\%) on DomainNet ($\lambda_O$ = 0.1)}. 
For the case where inputs have known domain (KD) label, we can use the corresponding learning mask (\model-KD) to achieve the strongest performance without requiring additional models or parameters. Column headers identify
the target domains in the corresponding multi-source shifts.
\\ \footnotesize $^\mp$We were unable to optimize the MetaReg~\cite{balaji2018metareg} objective with Adam~\cite{kingma2014adam} as the optimizer and therefore, we also include comparisons with Aggregate and MetaReg trained with SGD.}
\resizebox{0.8\columnwidth}{!}{
\begin{tabular}{l l l  c  c c c c c c | c }
\toprule
& & \textbf{Method} & & C & I & P & Q & R & S & Overall\\
\midrule
\multirow{7}{*}{\rotatebox{90}{\centering AlexNet }} & 
& Aggregate && 48.56 & 57.24 & 51.38 & 49.60 & 47.48 & 50.72 & 50.83\\
& & Aggregate-SGD$^\mp$ && 48.14 & 54.93 & 50.55 & 48.33 & 47.57 & 49.98 & 49.92\\
& & Multi-Headed && 48.16 & 56.73 & 51.31 & 49.75 & 47.65 & 50.82 & 50.74 \\
& & MetaReg~\cite{balaji2018metareg}$^\mp$ && 48.87 & 56.06 & 51.23 & 49.60 & 48.66 &50.12 & 50.76\\
\cline{3-11}
& & \model (Ours)  && 49.63 & 58.47 & 52.88 & 51.33 & 49.07 & 52.42 & 52.30 \\
& & \model-KD (Ours)  && \textbf{51.91}  & \textbf{61.01}  & \textbf{54.93}  & \textbf{53.84}  & \textbf{51.08}  & \textbf{54.47}  & \textbf{54.54}  \\
\midrule
\multirow{6}{*}{\rotatebox{90}{\centering ResNet-18 }} & 
& Aggregate && 56.58 & 65.27 & 59.29 & 59.15 & 55.47 & 58.84 & 59.10\\
& & Aggregate-SGD$^\mp$ && 55.32 & 63.63 & 57.40 & 57.98 & 53.99 & 57.37 & 57.62\\
& & Multi-Headed && 47.79 & 56.80 & 50.85 & 54.86 & 46.92 & 49.50 & 51.12 \\
& & MetaReg~\cite{balaji2018metareg}$^\mp$ && 56.25 & 63.07 & 57.74 & 58.73 & 55.40 & 58.04 & 58.21\\
\cline{3-11}
& & \model (Ours)  && 57.39 & 65.73 & 58.87 & 59.66 & 55.95 & 58.63 & 59.37 \\
& & \model-KD (Ours)  && \textbf{58.61}  & \textbf{66.98}  & \textbf{59.86}  & \textbf{60.98}  & \textbf{57.24}  & \textbf{59.84}  & \textbf{60.59}  \\
\midrule
\multirow{6}{*}{\rotatebox{90}{\centering ResNet-50 }} & 
& Aggregate && 61.68 & 69.73 & 63.90 & 63.88 & 60.29 & 63.62 & 63.85\\
& & Aggregate-SGD$^\mp$ && 61.64 & 69.36 & 63.65 & 64.08 & 60.52 & 63.82 & 63.85\\
& & Multi-Headed && 53.77 & 62.09 & 56.54 & 60.32 & 51.38 & 55.10 & 56.53 \\
& & MetaReg~\cite{balaji2018metareg}$^\mp$ && 61.86 & 68.80 & 63.23 & 64.75 & 60.59 & 63.21 & 63.74\\
\cline{3-11}
& & \model (Ours)  && 61.78 & 69.49 & 63.93 & 64.09 & 59.92 & 63.50 & 63.79 \\
& & \model-KD (Ours)  && \textbf{63.16}  & \textbf{70.79}  & \textbf{65.03}  & \textbf{65.67}  & \textbf{61.30}  & \textbf{64.86}  & \textbf{65.14}  \\
\bottomrule
\end{tabular}
}
\label{tab:keyidres_dmnt}
\end{center}
\end{table}

\begin{table}[t] \footnotesize
\renewcommand*{\arraystretch}{1.18}
\setlength{\tabcolsep}{6pt}
\begin{center}
\caption{\textbf{Domain-Specialized Masks ($\lambda_O$ = 0.1)}. We show how optimizing for
\texttt{sIoU} leads to masks which are specialized for the individual source
domains in terms of predictive performance. We consider two multi-source shifts I,P,Q,R,S$\rightarrow$C [top-half] and C,I,P,R,S$\rightarrow$Q [bottom-half] on DomainNet~\cite{peng2019moment} with the AlexNet as the backbone architecture and find that using corresponding source domain
masks leads to significantly improved in-domain performance.
}
\resizebox{0.7\columnwidth}{!}{
\begin{tabular}{l l l  c  c c c c c | c }
\toprule
& &  & &  &  & \textbf{Source} &  &  & \textbf{Target} \\
\midrule
& & \textbf{Chosen Mask} &  & I & P & Q & R & S & C \\
\midrule
\multirow{6}{*}{\rotatebox{90}{\centering AlexNet }} & 
& $\mathbf{m}^{\text{Infograph}}$ && \cellcolor{lightgray}{\textbf{23.84}}  & 45.56  & 59.13  & 62.43  & 46.70  & 46.91\\
& & $\mathbf{m}^{\text{Painting}}$ && 19.88  & {\cellcolor{lightgray}{\textbf{52.41}}} & 59.00 & 60.36 & 45.75 & 46.87\\
& & $\mathbf{m}^{\text{Quickdraw}}$ && 21.72 & 48.47 & {\cellcolor{lightgray}{\textbf{62.52}}} & 65.32 & 48.69 & 50.33\\
& & $\mathbf{m}^{\text{Real}}$ && 18.42 & 43.48 & 58.80 & {\cellcolor{lightgray}{\textbf{68.62}}} & 44.81 & 47.69\\
& & $\mathbf{m}^{\text{Sketch}}$ && 19.45 & 45.41 & 57.64 & 61.78 & {\cellcolor{lightgray}{\textbf{52.16}}} & 48.36\\
\cline{3-10}
& & Combined && 22.28 & 49.55 & 60.45 & 66.14 & 49.72 & 50.06\\
\midrule
& & \textbf{Chosen Mask} & & C & I & P & R & S & Q \\
\midrule
\multirow{6}{*}{\rotatebox{90}{\centering AlexNet }} & 
& $\mathbf{m}^{\text{Clipart}}$ && {\cellcolor{lightgray}{\textbf{66.70}}} & 21.36 & 46.60 & 64.35 & 49.70 & 13.37\\
& & $\mathbf{m}^{\text{Infograph}}$ && 60.71 & {\cellcolor{lightgray}{\textbf{24.95}}} & 47.06 & 63.78 & 49.36 & 12.58\\
& & $\mathbf{m}^{\text{Painting}}$ && 59.21 & 20.59 & {\cellcolor{lightgray}{\textbf{53.21}}} & 60.67 & 48.14 & 12.01\\
& & $\mathbf{m}^{\text{Real}}$ && 59.62 & 19.41 & 43.82 & {\cellcolor{lightgray}{\textbf{69.82}}} & 47.22 & 11.31\\
& & $\mathbf{m}^{\text{Sketch}}$ && 60.97 & 20.29 & 45.69 & 62.40 & {\cellcolor{lightgray}{\textbf{54.51}}} & 13.08\\
\cline{3-10}
& & Combined && 64.13 & 23.21 & 50.05 & 67.03 & 52.24 & 13.07\\
\bottomrule
\end{tabular}}
\label{tab:dmsp_masks_dmnt}
\end{center}
\end{table}
\subsection{Results}
We report results on both PACS (out-of-domain) and DomainNet (in-domain and out-of-domain). 
For DomainNet, we use C, I, P, Q, R, S to denote the domains -- \textit{clipart}, \textit{infograph}, \textit{painting}, \textit{quickdraw}, \textit{real} and \textit{sketch} respectively. On PACS, we use A, C, P and S to denote the domains -- \textit{art-painting}, \textit{cartoon}, \textit{photo} and \textit{sketch} respectively. We summarize the observed trends below:

\par \noindent
\textbf{Out-of-Domain Generalization.} Tables~\ref{tab:keyodres_dmnt} and~\ref{tab:keyodres_pacs}\footnote{For more comparisons to prior work, please refer to the appendix.}
    summarize out of domain generalization results on the DomainNet and PACS datasets, respectively.
    \par
    \textbf{DomainNet -} On DomainNet, we observe that \model beats the naive aggregate baseline, the multi-headed baseline and
    MetaReg~\cite{balaji2018metareg} 
    using AlexNet as the backbone architecture in terms of overall performance -- with an improvement of 
    2.7\% over MetaReg~\cite{balaji2018metareg} and 2.6\% over the Aggregate baseline. Interestingly,  this corresponds to an almost
    2.89\% improvement on the I,P,Q,R,S$\rightarrow$C and a 2.63\% improvement on the C,I,P,Q,S$\rightarrow$R shifts (see Table~\ref{tab:keyodres_dmnt}, AlexNet set of rows). 
    Using ResNet-18 as the backbone architecture, we observe that \model is competitive with
    MetaReg~\cite{balaji2018metareg} (improvement margin of 0.43\%) accompanied
    by improvements on the I,P,Q,R,S$\rightarrow$C and C,I,P,R,S$\rightarrow$Q shifts.
    We observe similar trends using ResNet-50, where \model is competitive with the
    best performing Aggregate-SGD$^\mp$ baseline.

    \textbf{PACS -} To compare \model with prior
    work in the Domain Generalization literature, we also report results on the more commonly used 
    PACS~\cite{Li2017dg} benchmark in Table~\ref{tab:keyodres_pacs}. We find that in terms of overall performance, \model with AlexNet as the backbone architecture outperforms
    baselines and prior approaches including
    MetaReg~\cite{balaji2018metareg}\footnote{We report the performance for MetaReg~\cite{balaji2018metareg} from \cite{li2019episodic} as the official PACS \texttt{train}-\texttt{val} data split changed post MetaReg~\cite{balaji2018metareg} publication.} -- which learns regularizers by modeling domain-shifts within the source set of distributions, MLDG~\cite{li2018learning} -- which learns robust network parameters using meta-learning and Epi-FCR~\cite{li2019episodic} -- a recently proposed episodic scheme to learn network parameters
    robust to domain-shift, 
    and performs competitively with MASF~\cite{dou2019domain} --  which introduces complementary losses to explicitly regularize the semantic structure of the
feature space via a model-agnostic episodic learning procedure. Notice that this improvement also comes with a 4.09\% improvement
    over MASF~\cite{dou2019domain} on the A,C,P$\rightarrow$S shift.
    Using ResNet-18 and ResNet-50 as the backbone architectures, we observe that \model leads to comparable and improved overall performance, with margins of 0.04\% and 0.7\% for ResNet-18 and ResNet-50, respectively. 
    For ResNet-18, this is accompanied with a 0.91\% and 1.92\% improvement on the
    A,C,P$\rightarrow$S and A,P,S$\rightarrow$C shifts. Similarily for ResNet-50, we observe a 3.06\% improvement on the A,C,P$\rightarrow$S shift. 
    
    Due to its increased size, both in terms of number of images and number of categories, DomainNet proves to be a more challenging benchmark than PACS. 
    Likely due to this difficulty, we find that performance on some of the hardest shifts
    (with Quickdraw and Infograph as the target domain) is significantly low ($<$25\% for Quickdraw). Furthermore,
    \model and prior domain generalization approaches perform comparably to naive baselines (ex. Aggregate) on these shifts, indicating that there is significant room for improvement.

\begin{figure}[t]
\centering
\includegraphics[width=0.8\textwidth]{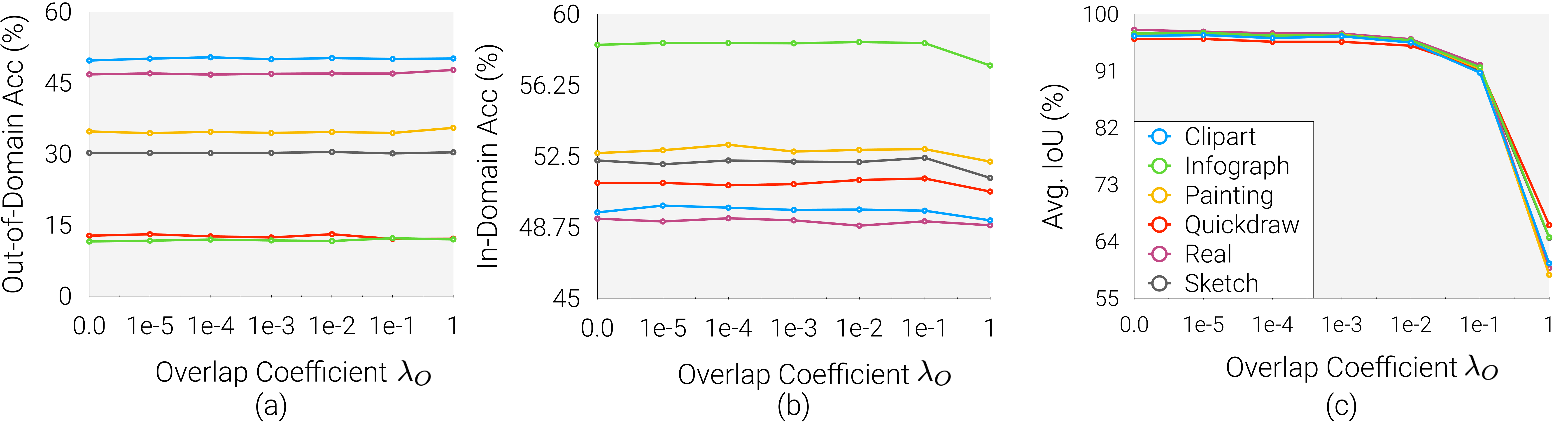}
\caption{\textbf{Sensitivity to $\lambda_O$.} \model is relatively insensitive to the setting of the hyper-parameter $\lambda_O$ as measured by  out-of-domain accuracy (a), in-domain accuracy (b), and average IoU score measured among pairs of source domain masks (c). The legends in
    (c) indicate the target domain in the corresponding multi-source
    shift. AlexNet is the backbone CNN.
    }
\label{fig:ov_ap}
\end{figure}

\par \noindent
\textbf{In-Domain Generalization.} For each of the domain-shifts in Table.~\ref{tab:keyodres_dmnt}, we further report in-domain generalization performance on DomainNet
    in Table.~\ref{tab:keyidres_dmnt}. 
    For in-domain evaluation, we present both our standard approach as well as a version which assumes knowledge of the domain corresponding to each test instance. 
    For the latter, we 
    report the performance of \model using only the mask corresponding to the known domain (KD) label and refer to this as 
    \model-KD.
    Notably, for this case where a test instance is drawn from one of the source domains, 
    \model-KD
    provides significant performance improvement over the baselines
    (see Table~\ref{tab:keyidres_dmnt}). Compared to \model, we observe that 
    \model-KD
    results in a consistent improvement of $\sim$1-2\%.
    This alludes to the fact that the learnt domain-specific masks are indeed
    specialized for individual source domains.

\section{Analysis}
\label{sec:analysis}
\par \noindent
\textbf{Domain Specialization.} \prithvi{
We demonstrate that as an outcome of \model, using masks 
corresponding to the source domain at hand leads to siginificantly improved
in-domain performance compared to a mismatched domain-mask pair, indicating the emergence of domain-specialized masks.}
\prithvi{In Table.~\ref{tab:dmsp_masks_dmnt}, we report results on the
I,P,Q,R,S$\rightarrow$C (easy) and C,I,P,R,S$\rightarrow$Q (hard) shifts using AlexNet as the
backbone CNN.
We report both in and out-of-domain performance 
using each of the source domain masks and compare it with the setting
when predictions from all the source domain masks are averaged. The cells
highlighted in gray represent in-domain accuracies when masks are paired
with the corresponding source domain. Clearly, using the mask corresponding
to the source domain instance at test-time (also see \model-KD in Table. 3) leads to significantly 
improved performance compared to the mis-matched pairs -- with differences with the second best source domain mask ranging from $\sim$2-4\% for 
I,P,Q,R,S$\rightarrow$C and $\sim$3-6\% for C,I,P,R,S$\rightarrow$Q.
This indicates that not only do the source domain masks overlap
minimally, but they are also ``specialized'' for each of the source
domains in terms of predictive performance.
We further observe that averaging predictions obtained from all the
source domain masks leads to performance that is relatively closer to the \model-KD
setting compared to a mismatched mask-domain pair (but still falls behind
by $\sim$2-3\%).}
We note that certain source domain masks do lead
to out-of-domain accuracies which are close (within 1\%) to the combined setting -- $\mathbf{m}^{\text{Quickdraw}}$ for the I,P,Q,R,S$\rightarrow$C shift and $\mathbf{m}^{\text{Clipart}}$, $\mathbf{m}^{\text{Infograph}}$, $\mathbf{m}^{\text{Sketch}}$ for the C,I,P,R,S$\rightarrow$Q shift. This 
highlights the motivation
at the heart of our approach -- how leveraging
characteristics specific to individual source domains in addition to
the invariant ones 
are useful for generalization.
\par \noindent
\textbf{Sensitivity to $\lambda_O$.}
A key component of our approach is the 
\textit{soft-}IoU loss which encourages domain specificity by minimizing overlapping features across domains.
During optimization, we require setting of a loss balancing hyper-parameter, $\lambda_O$. Here, we explore the sensitivity of our model to 
$\lambda_O$
by sweeping from 0 to 1 in logarithmic increments. 
Fig.~\ref{fig:ov_ap}
shows the final in and out-of-domain accuracies (Fig.~\ref{fig:ov_ap} (b) and (a))
and overlap (Fig.~\ref{fig:ov_ap} (c)) measured 
as the IoU~\cite{article} among pairs of \textit{discrete} source domain masks obtained by thresholding the
soft-mask values per-domain at 0.5, i.e., $m = \mathbf{1}_{\mathbf{m}^d > 0.5}$ for domain $d$.
We observe that both 
in and out-of-domain
generalization performance is robust to the choice of $\lambda_O$, with only minor variations
and a slight drop in in-domain performance at extreme values of $\lambda_O$ (0.1 and 1). In Fig.~\ref{fig:ov_ap} (c), we observe that initially average pairwise IoU measures stay stable
till $\lambda_O = 10^{-3}$ but drop at high values of $\lambda_O = 0.1 \text{ and }1$ (as low
as $<60\%$ for some shifts)-- indicating
an increase in the ``domain specificity'' of the masks involved. 
Note
that low IoU
at high-values of $\lambda_O$ 
is accompanied only by
a minor drop in in-domain performance and almost 
no-drop in out-of-domain performance! 
It is crucial to note here that although there is an expected 
trade-off between specificity and generalization performance this trade-off does not result in large fluctuations for \model. Please refer to the appendix for more analysis of \model.

\section{Conclusion}
To summarize, we propose \model : \textbf{D}omain-specific
\textbf{M}asks for \textbf{G}eneralization, a method for multi-source domain learning which
balances domain-specific and domain-invariant feature representations to produce a single strong model capable of effective domain generalization.
We learn this balance by introducing domain-specific masks over neurons and optimizing such masks so as to minimize cross-domain feature overlap. 
Thus, our model, \model, benefits from the predictive power of
features specific to individual domains while retaining the generalization capapbilities of components shared across the source domains. \model achieves competitive out-of-domain performance on the 
commonly used PACS dataset and competitive in and out-of-domain performance on the challenging DomainNet dataset. Although beyond the scope of this paper, encouraging
a blend of domain specificity and invariance may be useful not only in the
context of generalization performance but also in terms of model interpretability.
\par \noindent
\textbf{Acknowledgements.} We thank Viraj Prabhu, Daniel Bolya, Harsh Agrawal and Ramprasaath Selvaraju for fruitful discussions and feedback. This work was partially supported by DARPA award FA8750-19-1-0504. 

\bibliographystyle{splncs04}
\bibliography{main}

\begin{thebibliography}{10}
\providecommand{\url}[1]{\texttt{#1}}
\providecommand{\urlprefix}{URL }
\providecommand{\doi}[1]{https://doi.org/#1}

\bibitem{ahuja2020invariant}
Ahuja, K., Shanmugam, K., Varshney, K., Dhurandhar, A.: Invariant risk
  minimization games. arXiv preprint arXiv:2002.04692  (2020)

\bibitem{arjovsky2019invariant}
Arjovsky, M., Bottou, L., Gulrajani, I., Lopez-Paz, D.: Invariant risk
  minimization. arXiv preprint arXiv:1907.02893  (2019)

\bibitem{balaji2018metareg}
Balaji, Y., Sankaranarayanan, S., Chellappa, R.: Metareg: Towards domain
  generalization using meta-regularization. In: Advances in Neural Information
  Processing Systems. pp. 998--1008 (2018)

\bibitem{bengio2013estimating}
Bengio, Y., L{\'e}onard, N., Courville, A.: Estimating or propagating gradients
  through stochastic neurons for conditional computation. arXiv preprint
  arXiv:1308.3432  (2013)

\bibitem{berriel2019budget}
Berriel, R., Lathuillere, S., Nabi, M., Klein, T., Oliveira-Santos, T., Sebe,
  N., Ricci, E.: Budget-aware adapters for multi-domain learning. In:
  Proceedings of the IEEE International Conference on Computer Vision. pp.
  382--391 (2019)

\bibitem{Bousmalis_2017_CVPR}
Bousmalis, K., Silberman, N., Dohan, D., Erhan, D., Krishnan, D.: Unsupervised
  pixel-level domain adaptation with generative adversarial networks. In: The
  IEEE Conference on Computer Vision and Pattern Recognition (CVPR) (July 2017)

\bibitem{bousmalis2016domain}
Bousmalis, K., Trigeorgis, G., Silberman, N., Krishnan, D., Erhan, D.: Domain
  separation networks. In: Advances in neural information processing systems.
  pp. 343--351 (2016)

\bibitem{burgess2018understanding}
Burgess, C.P., Higgins, I., Pal, A., Matthey, L., Watters, N., Desjardins, G.,
  Lerchner, A.: Understanding disentangling in $\beta$-vae. arXiv preprint
  arXiv:1804.03599  (2018)

\bibitem{NIPS2016_6399}
Chen, X., Duan, Y., Houthooft, R., Schulman, J., Sutskever, I., Abbeel, P.:
  Infogan: Interpretable representation learning by information maximizing
  generative adversarial nets. In: Advances in neural information processing
  systems. pp. 2172--2180 (2016)

\bibitem{deng2009imagenet}
Deng, J., Dong, W., Socher, R., Li, L.J., Li, K., Fei-Fei, L.: Imagenet: A
  large-scale hierarchical image database. In: 2009 IEEE conference on computer
  vision and pattern recognition. pp. 248--255. Ieee (2009)

\bibitem{dou2019domain}
Dou, Q., de~Castro, D.C., Kamnitsas, K., Glocker, B.: Domain generalization via
  model-agnostic learning of semantic features. In: Advances in Neural
  Information Processing Systems. pp. 6447--6458 (2019)

\bibitem{finn2017model}
Finn, C., Abbeel, P., Levine, S.: Model-agnostic meta-learning for fast
  adaptation of deep networks. In: Proceedings of the 34th International
  Conference on Machine Learning-Volume 70. pp. 1126--1135. JMLR. org (2017)

\bibitem{french2018selfensembling}
French, G., Mackiewicz, M., Fisher, M.: Self-ensembling for visual domain
  adaptation. In: International Conference on Learning Representations (2018),
  \url{https://openreview.net/forum?id=rkpoTaxA-}

\bibitem{ganin2016domain}
Ganin, Y., Ustinova, E., Ajakan, H., Germain, P., Larochelle, H., Laviolette,
  F., Marchand, M., Lempitsky, V.: Domain-adversarial training of neural
  networks. The Journal of Machine Learning Research  \textbf{17}(1),
  2096--2030 (2016)

\bibitem{ghifary2015domain}
Ghifary, M., Bastiaan~Kleijn, W., Zhang, M., Balduzzi, D.: Domain
  generalization for object recognition with multi-task autoencoders. In:
  Proceedings of the IEEE international conference on computer vision. pp.
  2551--2559 (2015)

\bibitem{MTAE}
Ghifary, M., Kleijn, W.B., Zhang, M., Balduzzi, D.: Domain generalization for
  object recognition with multi-task autoencoders. In: 2015 {IEEE}
  International Conference on Computer Vision, {ICCV} 2015, Santiago, Chile,
  December 7-13, 2015 (2015)

\bibitem{he2016deep}
He, K., Zhang, X., Ren, S., Sun, J.: Deep residual learning for image
  recognition. In: Proceedings of the IEEE conference on computer vision and
  pattern recognition. pp. 770--778 (2016)

\bibitem{hoffman2018cycada}
Hoffman, J., Tzeng, E., Park, T., Zhu, J., Isola, P., Saenko, K., Efros, A.A.,
  Darrell, T.: Cycada: Cycle-consistent adversarial domain adaptation. In:
  Proceedings of the 35th International Conference on Machine Learning, {ICML}
  2018, Stockholmsm{\"{a}}ssan, Stockholm, Sweden, July 10-15, 2018. pp.
  1994--2003 (2018)

\bibitem{article}
Jaccard, P.: Etude de la distribution florale dans une portion des alpes et du
  jura. Bulletin de la Societe Vaudoise des Sciences Naturelles  \textbf{37},
  547--579 (01 1901). \doi{10.5169/seals-266450}

\bibitem{khosla2012undoing}
Khosla, A., Zhou, T., Malisiewicz, T., Efros, A.A., Torralba, A.: Undoing the
  damage of dataset bias. In: European Conference on Computer Vision. pp.
  158--171. Springer (2012)

\bibitem{Undo_bias}
Khosla, A., Zhou, T., Malisiewicz, T., Efros, A.A., Torralba, A.: Undoing the
  damage of dataset bias. In: European Conference on Computer Vision. pp.
  158--171. Springer (2012)

\bibitem{pmlr-v80-kim18b}
Kim, H., Mnih, A.: Disentangling by factorising. In: Dy, J., Krause, A. (eds.)
  Proceedings of the 35th International Conference on Machine Learning.
  Proceedings of Machine Learning Research, vol.~80, pp. 2649--2658. PMLR,
  Stockholmsmässan, Stockholm Sweden (10--15 Jul 2018)

\bibitem{kingma2014adam}
Kingma, D.P., Ba, J.: Adam: A method for stochastic optimization. arXiv
  preprint arXiv:1412.6980  (2014)

\bibitem{krizhevsky2012imagenet}
Krizhevsky, A., Sutskever, I., Hinton, G.E.: Imagenet classification with deep
  convolutional neural networks. In: Advances in neural information processing
  systems. pp. 1097--1105 (2012)

\bibitem{lee2019drop}
Lee, S., Kim, D., Kim, N., Jeong, S.G.: Drop to adapt: Learning discriminative
  features for unsupervised domain adaptation. In: Proceedings of the IEEE
  International Conference on Computer Vision. pp. 91--100 (2019)

\bibitem{Li2017dg}
Li, D., Yang, Y., Song, Y.Z., Hospedales, T.: Deeper, broader and artier domain
  generalization. In: International Conference on Computer Vision (2017)

\bibitem{li2017deeper}
Li, D., Yang, Y., Song, Y.Z., Hospedales, T.M.: Deeper, broader and artier
  domain generalization. In: Proceedings of the IEEE International Conference
  on Computer Vision. pp. 5542--5550 (2017)

\bibitem{li2018learning}
Li, D., Yang, Y., Song, Y.Z., Hospedales, T.M.: Learning to generalize:
  Meta-learning for domain generalization. In: Thirty-Second AAAI Conference on
  Artificial Intelligence (2018)

\bibitem{li2019episodic}
Li, D., Zhang, J., Yang, Y., Liu, C., Song, Y.Z., Hospedales, T.M.: Episodic
  training for domain generalization. In: Proceedings of the IEEE International
  Conference on Computer Vision. pp. 1446--1455 (2019)

\bibitem{long2015DAN}
Long, M., Cao, Y., Wang, J., Jordan, M.I.: Learning transferable features with
  deep adaptation networks. In: Proceedings of the 32nd International
  Conference on Machine Learning. pp. 97--105 (2015)

\bibitem{long2016RTN}
Long, M., Wang, J., Jordan, M.I.: Unsupervised domain adaptation with residual
  transfer networks. CoRR  \textbf{abs/1602.04433} (2016)

\bibitem{Long2017JAN}
Long, M., Zhu, H., Wang, J., Jordan, M.I.: Deep transfer learning with joint
  adaptation networks. In: Precup, D., Teh, Y.W. (eds.) Proceedings of the 34th
  International Conference on Machine Learning, {ICML} 2017, Sydney, NSW,
  Australia, 6-11 August 2017. Proceedings of Machine Learning Research,
  vol.~70, pp. 2208--2217. {PMLR} (2017)

\bibitem{mallya2018piggyback}
Mallya, A., Davis, D., Lazebnik, S.: Piggyback: Adapting a single network to
  multiple tasks by learning to mask weights. In: Proceedings of the European
  Conference on Computer Vision (ECCV). pp. 67--82 (2018)

\bibitem{mallya2018packnet}
Mallya, A., Lazebnik, S.: Packnet: Adding multiple tasks to a single network by
  iterative pruning. In: Proceedings of the IEEE Conference on Computer Vision
  and Pattern Recognition. pp. 7765--7773 (2018)

\bibitem{mancini2018best}
Mancini, M., Bul{\`o}, S.R., Caputo, B., Ricci, E.: Best sources forward:
  domain generalization through source-specific nets. In: 2018 25th IEEE
  International Conference on Image Processing (ICIP). pp. 1353--1357. IEEE
  (2018)

\bibitem{muandet2013domain}
Muandet, K., Balduzzi, D., Sch{\"o}lkopf, B.: Domain generalization via
  invariant feature representation. In: International Conference on Machine
  Learning. pp. 10--18 (2013)

\bibitem{NEURIPS2019_9015}
Paszke, A., Gross, S., Massa, F., Lerer, A., Bradbury, J., Chanan, G., Killeen,
  T., Lin, Z., Gimelshein, N., Antiga, L., et~al.: Pytorch: An imperative
  style, high-performance deep learning library. In: Advances in neural
  information processing systems. pp. 8026--8037 (2019)

\bibitem{peng2019moment}
Peng, X., Bai, Q., Xia, X., Huang, Z., Saenko, K., Wang, B.: Moment matching
  for multi-source domain adaptation. In: Proceedings of the IEEE International
  Conference on Computer Vision. pp. 1406--1415 (2019)

\bibitem{Peng2019DomainAL}
Peng, X., Huang, Z., Sun, X., Saenko, K.: Domain agnostic learning with
  disentangled representations. In: ICML (2019)

\bibitem{rahman2016optimizing}
Rahman, M.A., Wang, Y.: Optimizing intersection-over-union in deep neural
  networks for image segmentation. In: International symposium on visual
  computing. pp. 234--244. Springer (2016)

\bibitem{saito2018adversarial}
Saito, K., Ushiku, Y., Harada, T., Saenko, K.: Adversarial dropout
  regularization. In: International Conference on Learning Representations
  (2018)

\bibitem{saito2017maximum}
Saito, K., Watanabe, K., Ushiku, Y., Harada, T.: Maximum classifier discrepancy
  for unsupervised domain adaptation. arXiv preprint arXiv:1712.02560  (2017)

\bibitem{sankaranarayanan2018GTA}
Sankaranarayanan, S., Balaji, Y., Castillo, C.D., Chellappa, R.: Generate to
  adapt: Aligning domains using generative adversarial networks. In: The IEEE
  Conference on Computer Vision and Pattern Recognition (CVPR) (June 2018)

\bibitem{savarese2019winning}
Savarese, P., Silva, H., Maire, M.: Winning the lottery with continuous
  sparsification. arXiv preprint arXiv:1912.04427  (2019)

\bibitem{serra2018overcoming}
Serra, J., Suris, D., Miron, M., Karatzoglou, A.: Overcoming catastrophic
  forgetting with hard attention to the task. In: International Conference on
  Machine Learning. pp. 4548--4557 (2018)

\bibitem{srivastava2014dropout}
Srivastava, N., Hinton, G., Krizhevsky, A., Sutskever, I., Salakhutdinov, R.:
  Dropout: a simple way to prevent neural networks from overfitting. The
  journal of machine learning research  \textbf{15}(1),  1929--1958 (2014)

\bibitem{tzeng2017adversarial}
Tzeng, E., Hoffman, J., Saenko, K., Darrell, T.: Adversarial discriminative
  domain adaptation. In: Proceedings of the IEEE conference on computer vision
  and pattern recognition. pp. 7167--7176 (2017)

\bibitem{venkatesh2020calibrate}
Venkatesh, B., Thiagarajan, J.J., Thopalli, K., Sattigeri, P.: Calibrate and
  prune: Improving reliability of lottery tickets through prediction
  calibration. arXiv preprint arXiv:2002.03875  (2020)

\bibitem{volpi2018generalizing}
Volpi, R., Namkoong, H., Sener, O., Duchi, J.C., Murino, V., Savarese, S.:
  Generalizing to unseen domains via adversarial data augmentation. In:
  Advances in Neural Information Processing Systems. pp. 5334--5344 (2018)

\bibitem{xu2018deep}
Xu, R., Chen, Z., Zuo, W., Yan, J., Lin, L.: Deep cocktail network:
  Multi-source unsupervised domain adaptation with category shift. In:
  Proceedings of the IEEE Conference on Computer Vision and Pattern
  Recognition. pp. 3964--3973 (2018)

\bibitem{zunino2020explainable}
Zunino, A., Bargal, S.A., Volpi, R., Sameki, M., Zhang, J., Sclaroff, S.,
  Murino, V., Saenko, K.: Explainable deep classification models for domain
  generalization. arXiv preprint arXiv:2003.06498  (2020)

\end{thebibliography}

\newpage

\section{Appendix}
In this appendix, we further discuss the specificity of the obtained domain-specific
masks (Section.~\ref{sec:domain_specialization}). Following this, we discuss how sparsity as an incentive
compares with \texttt{sIoU} in terms of learning a balance between specificity and invariance
and in terms of performance (Section.~\ref{sec:sp_vs_ov}). In Section.~\ref{sec:ens_test_time}, we discuss alternative techniques for directly ensembling masks instead of the output predictions in response to each mask. In Section.~\ref{sec:more_res}, we provide more extensive comparisons to prior work 
on the PACS~\cite{Li2017dg} dataset.
Finally, in Section.~\ref{sec:exp_details}, we describe in detail the
implementation and other details associated with our experiments. We use C, I, P, Q, R, S to denote the domains -- \textit{clipart}, \textit{infograph}, \textit{painting}, \textit{quickdraw}, \textit{real} and \textit{sketch} respectively on the DomainNet~\cite{peng2019moment} dataset.

\subsection{Domain Specificity}
\label{sec:domain_specialization}

As discussed in Section.~\ref{sec:feature_selection} (main paper), we incentivize domain specificity by optimizing the
\textit{soft-}IoU ($\mathtt{sIoU}$) objective (see Eqn.~\ref{eq:overall_obj} in main paper). To understand the extend of domain-specificity
achieved at convergence, we measure the Jaccard Similarity Coefficient~\cite{article} (also
known as IoU) among pairs of \textit{discrete} source domain masks, which we obtain by thresholding the
soft-mask values per-domain at 0.5, i.e., $m = \mathbf{1}_{\mathbf{m}^d > 0.5}$ for domain $d$.

\begin{figure}[t]
\centering
\includegraphics[width=0.8\columnwidth]{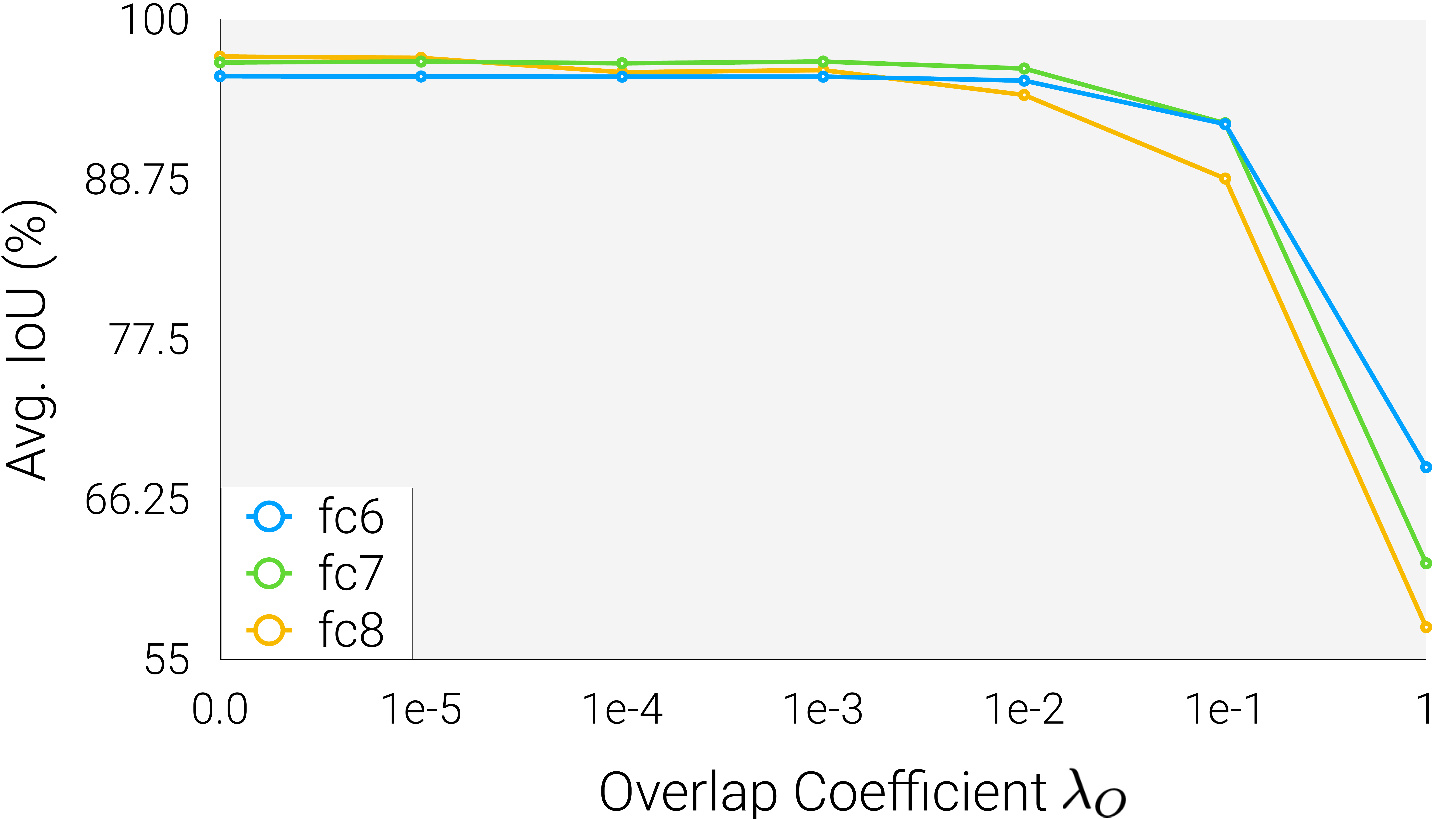}
\vspace{-7pt}
\caption{\textbf{Layerwise IoU sensitivity to $\lambda_O$.} The average IoU score among pairs of source domain masks decreases as $\lambda_O$ increases, indicating the degree to which domain-specificity emerges in individual layers (fc6, fc7, fc8). }
\label{fig:lyr_ov_sen}
\vspace{-20pt}
\end{figure}

Fig.~\ref{fig:ov_ap} shows the IoU among pairs of source domain
masks in addition to the overall average on DomainNet for the I,P,Q,R,S$\rightarrow$C and
C,I,P,R,S$\rightarrow$Q shifts with AlexNet as the backbone architecture ($\lambda_O$ = 0.1 during training). Note that the above metric provides information about the fraction of overlapping neurons which
are shared among pairs of source domains but only considers them among the ones which are activated (turned \textit{on}) based on the discrete masks $m$. Therefore, in addition to the IoU statistics
(as represented by the bars), we also report the fraction of activated neurons on average. 
We note that domain specificity
does emerge by learning masks in the manner described in Sec.~\ref{sec:feature_selection} of the main paper, as
evident by the IoU measures across pairs being lower than -- (1) $\sim$96\% for the
maximal pairwise IoU and (2) $\sim$92\% for overall IoU measures across both the shifts. Fig.~\ref{fig:lyr_ov_sen} shows how the layerwise overall IoU measure evolves
as $\lambda_O$ increases. While at lower values of $\lambda_O$, the amount of specificity is relatively low and similar across layers, at higher
values of $\lambda_O$ we see an increase of varying degrees across layers -- the relative ordering
among layers in terms of IoU being \texttt{fc6>fc7>fc8}, indicating the importance of
having \textit{more} shared neurons in the earlier layers.

\begin{figure*}[t]%
    \centering
    \subfloat[I,P,Q,R,S $\rightarrow$ C (O.O.D. Acc $50.06\%$)]{{\includegraphics[width=0.46\textwidth]{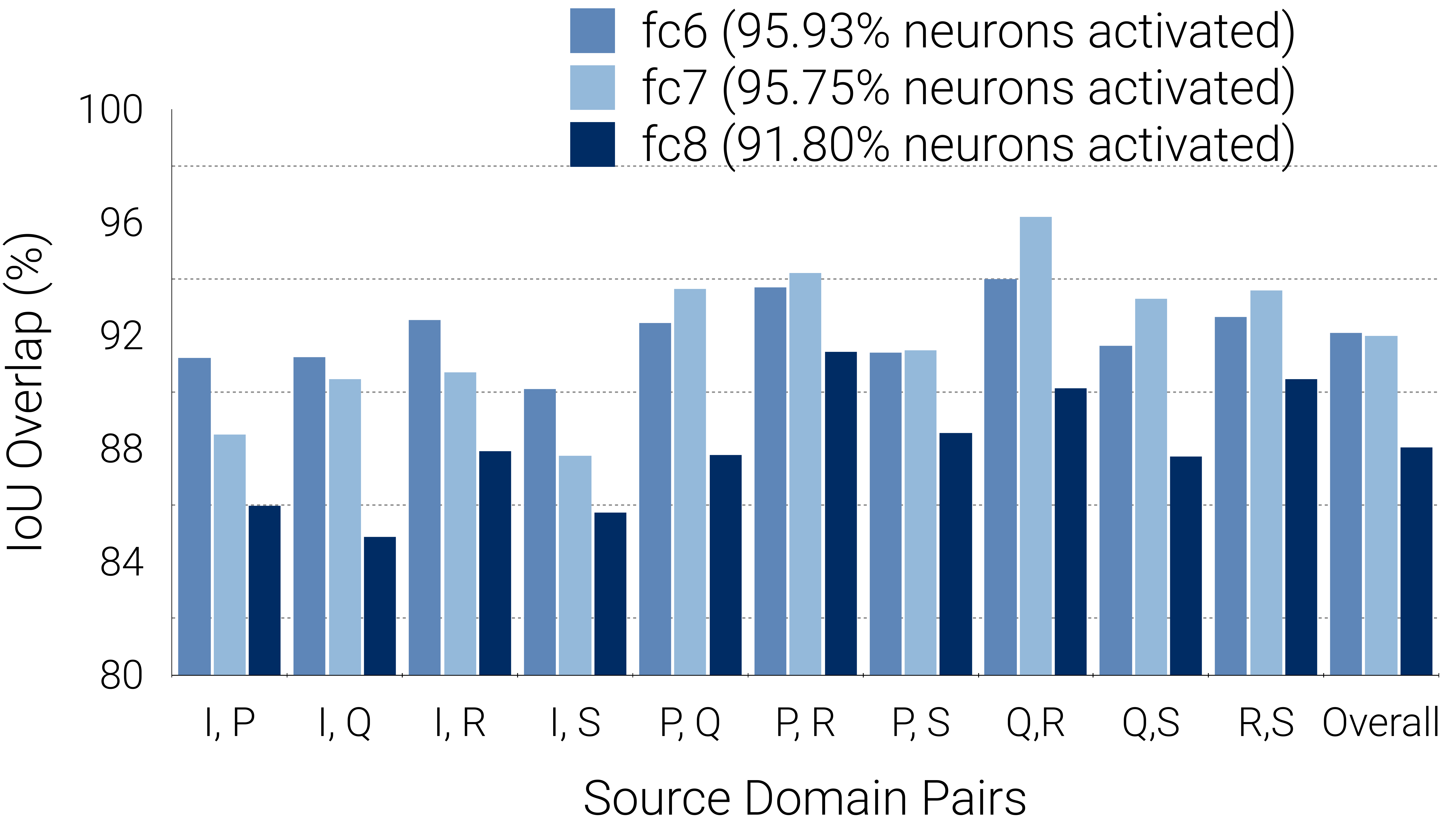} }}%
    \qquad
    \subfloat[C,I,P,R,S $\rightarrow$ Q (O.O.D. Acc $13.07\%$)]{{\includegraphics[width=0.46\textwidth]{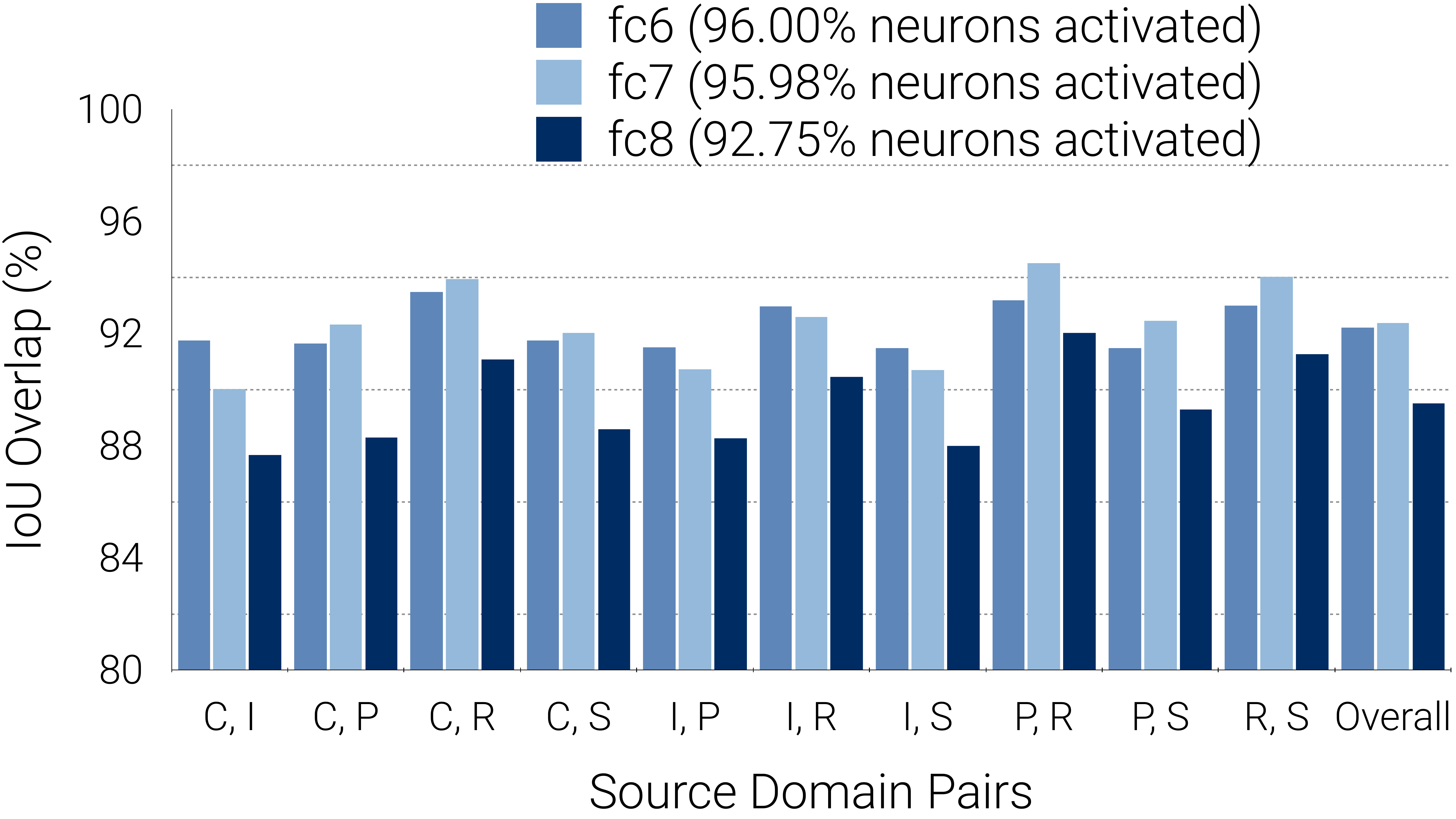} }}%
    \caption{\textbf{Emergence of domain-specificity in AlexNet with $\lambda_O=0.1$.} We show the IoU overlap among pairs of discrete source domain masks for the two shifts (a) I,P,Q,R,S$\rightarrow$C and (b) C,I,P,R,S $\rightarrow$Q on DomainNet~\cite{peng2019moment} with out-of-domain accuracies 48.70\% and 12.7\% respectively. We find that domain-specificity does indeed emerge, as indicated by the
    IoU measures.}
	\label{fig:ov_ap}
\end{figure*}

Finally, note that since the pairwise IoU
measures indicate the fraction of neurons which are shared among the neurons
which are turned \textit{on}, upon convergence we can essentially categorize the
neurons present in the task network into three categories -- (1) \textit{equally useless} -- neurons turned \textit{off} across all the source domain masks, (2) \textit{equally useful} or \textit{shared} -- neurons turned \textit{on} across all the source
domain masks and (3) \textit{domain-specific} -- neurons turned \textit{on} only
for specific source domains.

\subsection{Choice of Incentive: \texttt{sIoU} vs Sparsity}
\label{sec:sp_vs_ov}
\begin{figure*}[t]%
    \centering
    \subfloat[Out-of-Domain Acc]{{\includegraphics[width=0.32\textwidth]{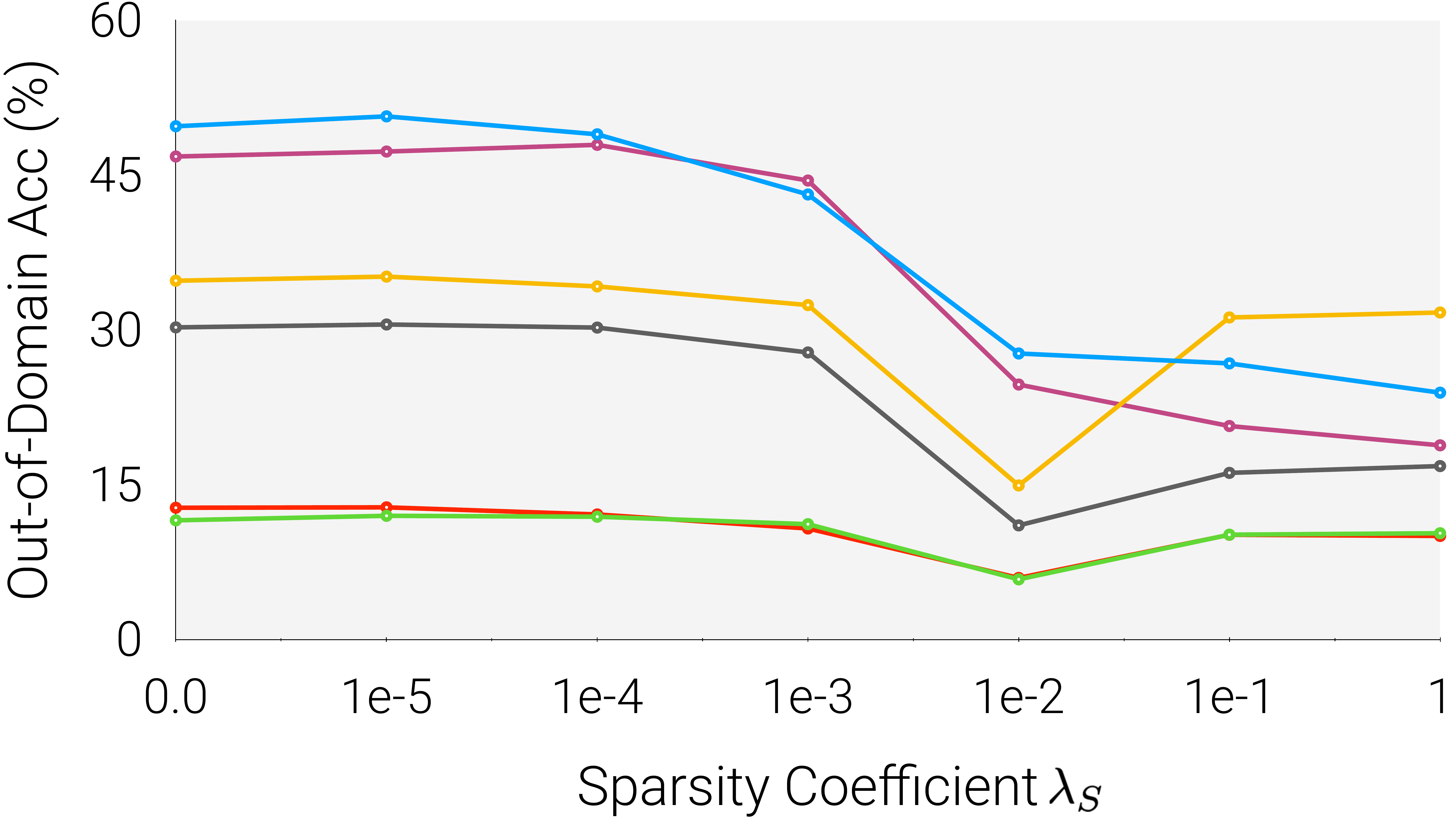} }}%
    \subfloat[In-Domain Acc]{{\includegraphics[width=0.32\textwidth]{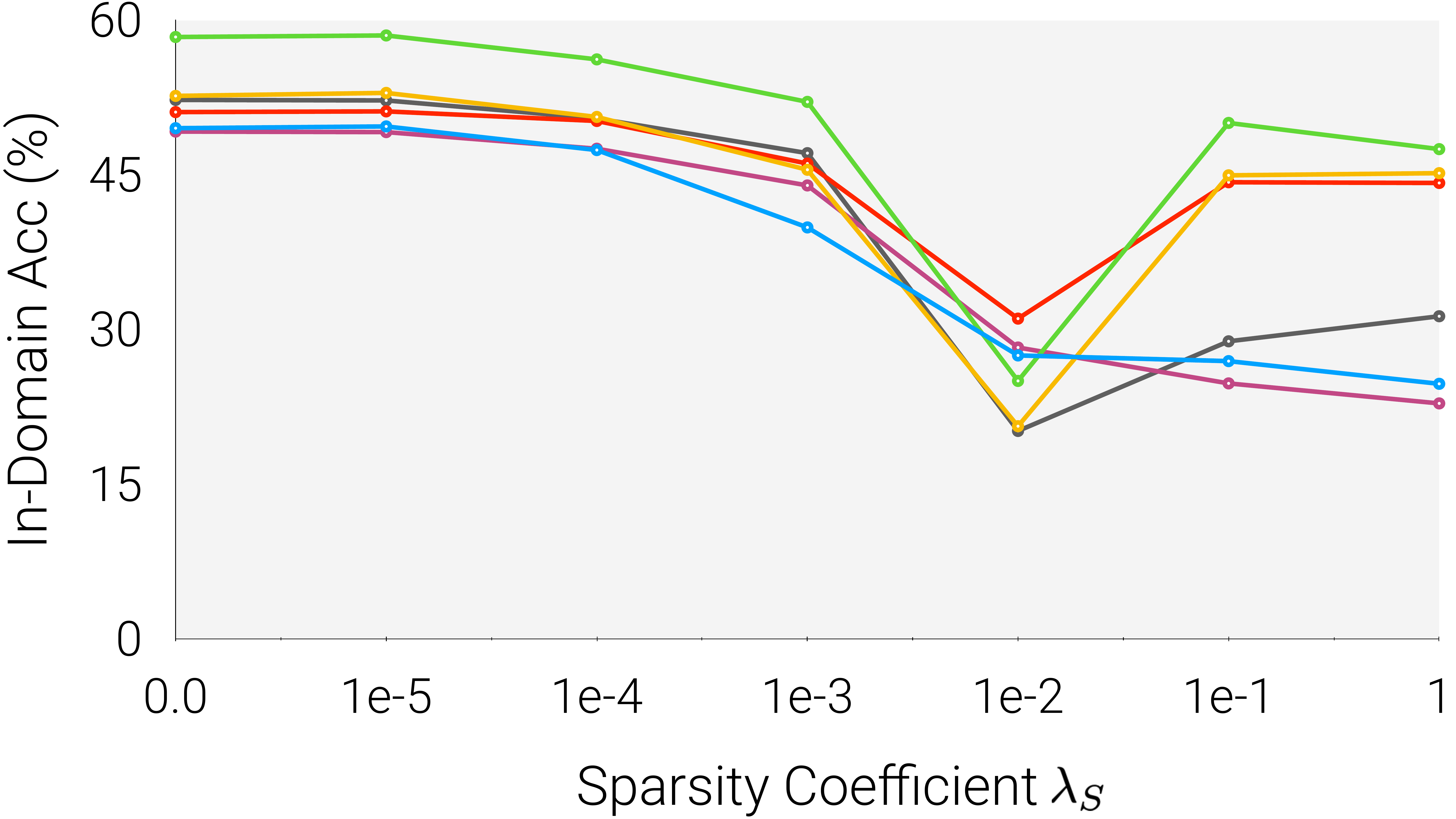} }}%
    \subfloat[Average IoU]{{\includegraphics[width=0.32\textwidth]{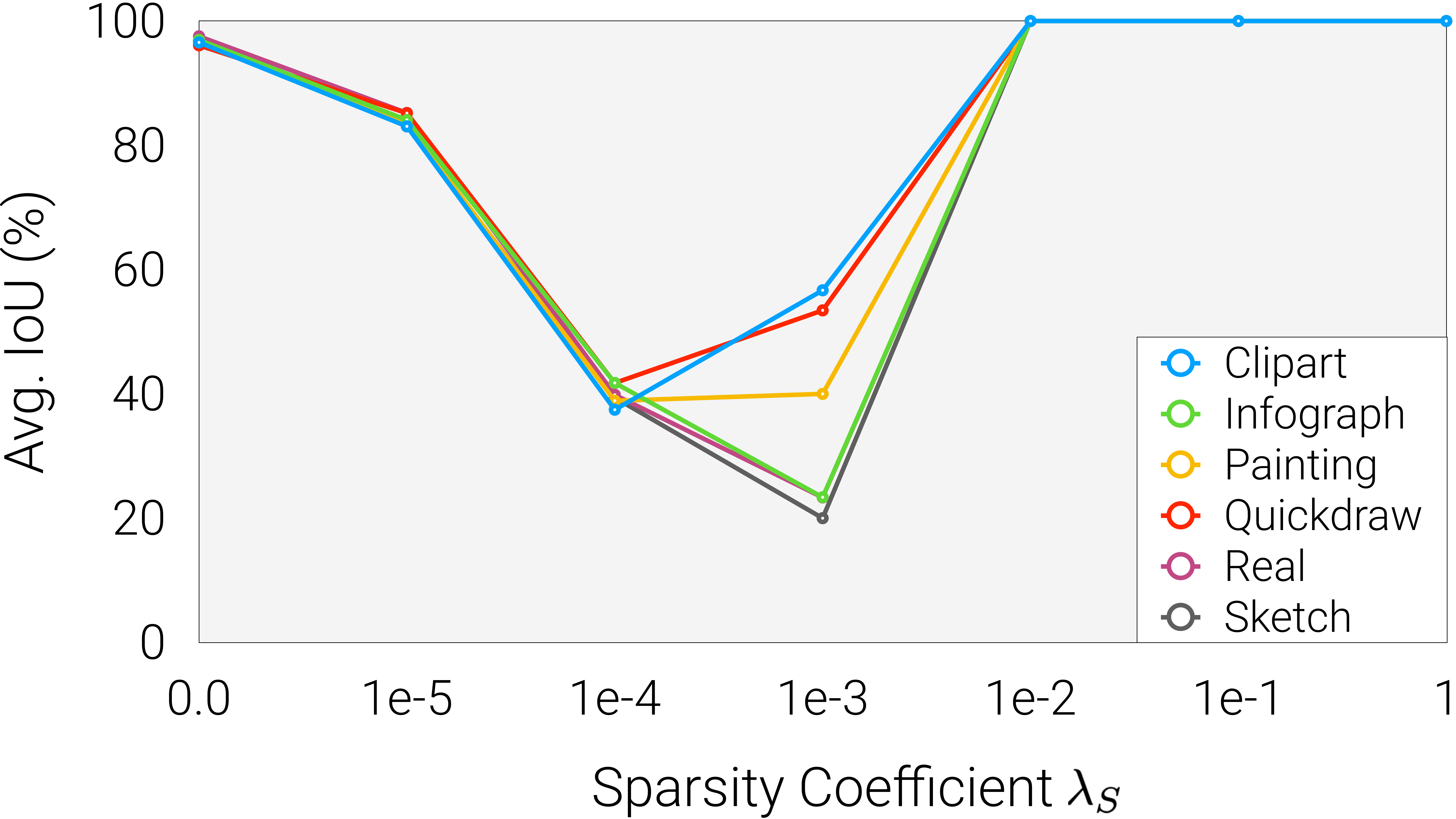} }}%
    \caption{\textbf{Sensitivity to $\lambda_S$.} We replace the \texttt{sIoU} with a differentiable sparsity term (coefficient $\lambda_S$) -- L1-norm of the \textit{soft-}source domain masks, i.e., $\sum_{D_i \in D_S} ||\mathbf{m}_i||_1$ -- and study the
    sensitivity to $\lambda_S$ as measured by out-of-domain accuracy (a), in-domain accuracy (b) and average IoU score measured among pairs of source domain masks. The legends in (b) indicate the target domain in
    the corresponding multi-source shift. We find that predictive
    performance and specificity (Avg. IoU) is \textit{very} sensitive to
    $\lambda_S$.}
	\label{fig:sp_ap}
\end{figure*}

As described in Section.~\ref{sec:feature_selection} (main paper), to ensure feature selection, we impose a \textit{soft-}IoU loss
in addition to standard cross-entropy training to penalize overlap among pairs of source
domain masks. However, in practice, one could also impose a sparsity constraint on the domain-sepcific masks
being learned ensure minimality in the number of features or neurons selected during learning. However, just incorporating a sparsity constraint does not explicitly incentivize domain-specificity -- masks corresponding to all the source domains could just end up picking the same
set of neurons, which is equivalent to learning a bottleneck layer during training.

We investigate the consequences of incorporating a sparsity regularizer in Figure.~\ref{fig:sp_ap}
on all the multi-source shifts of the DomainNet dataset using AlexNet as our backbone architecture.
Specifically, instead of the \texttt{sIoU} loss, we penalize the L1-norm of the soft-mask
values, i.e., $\mathbf{m}^d$ for all the source domains -- $\sum_{d \in D_S} ||\mathbf{m}^d||_1$\footnote{Since the soft-mask probabilities ($\mathbf{m}^d$) are positive, $||\mathbf{m}^d||_1$ is essentially the sum of mask probabilities per-neuron and is therefore differentiable and can be optimized using gradient descent.}.
We run a sweep over different values of the coefficient ($\lambda_S$) of this sparsity incentive from 0 to 1 in logarithmic increments. Fig.~\ref{fig:sp_ap} (a) and (b) show how out-of-domain and in-domain generalization performances and Fig.~\ref{fig:sp_ap} (b) shows how the pairwise IoU measure among the source
domain masks -- indicating domain-specificity, vary
with $\lambda_S$. Unlike $\lambda_O$ (see Sec.~\ref{sec:analysis}, main paper), we find that generalization performance is quite sensitive to the choice
of $\lambda_S$, with both out-of-domain and in-domain accuracies degrading significantly
at relatively high values of $\lambda_S$. We
find performance comparable to our approach only at values of $\lambda=10^{-5}$. For the pairwise IoU
measures, we observe that while specificity increases to some extent till $\lambda_S=10^{-3}$, but decreases sharply with further increase in $\lambda_S$. At high-values of $\lambda_S$, we observe that the source domain masks are extremely sparse and have high overlap
indicating the fact that the masks essentially encourage learning just a bottleneck layer.
This further demonstrates the efficacy of the \texttt{sIoU} loss in maintaining a reasonable
balance between encouraging specificty while retaining predictive performance.

\subsection{Ensembling Choices at Test-time}
\label{sec:ens_test_time}
In Section.~\ref{sec:feature_selection} (main paper), we describe how we follow a soft-scaling
scheme akin to dropout~\cite{srivastava2014dropout} at test-time. Specifically,
we obtain predictions corresponding to neurons in the task network
soft-scaled by individual source domain masks and average them (call this \texttt{Pred-Ens}). In this section, we further investigate if the choice
of ensembling method at test-time matters. We compare \texttt{Pred-Ens}
with the setting where we average the soft masks ($\mathbf{m}^d$ for source domain $d$) and draw a single prediction by scaling neurons with the
averaged soft-mask -- \texttt{Mask-Ens}.

In Table.~\ref{tab:ens_res_dmnt}, we compare \model (\texttt{Pred-Ens})
and \model (\texttt{Mask-Ens}) in terms of both in and out-of-domain performances
on all the multi-source shifts on DomainNet
using AlexNet as the backbone architecture. We observe that 
\texttt{Mask-Ens} performs comparatively with \texttt{Pred-Ens},
with the margin of difference being within $\sim$1\%.

\begin{table}[t]
\renewcommand*{\arraystretch}{1.18}
\setlength{\tabcolsep}{4pt}
\begin{center}
\resizebox{\columnwidth}{!}{
\begin{tabular}{l l l  c  c c c c c c | c }
\toprule
& & \textbf{Method} & & Clipart & Infograph & Painting & Quickdraw & Real & Sketch & Overall \\
\midrule
& & \textbf{Out-of-Domain} & & & & & & & & \\
\midrule
\multirow{6}{*}{\rotatebox{90}{\centering AlexNet }} & 
 & Aggregate && 47.17 & 10.15 & 31.82 & 11.75 & 44.35 & 26.33& 28.60\\
& & Aggregate-SGD$^\mp$ && 42.30 & 12.42 & 31.45 & 9.52 & 42.76 & 29.34& 27.97\\
& & Multi-Headed  && 45.96 & 10.56 & 31.07 & 12.05 & 43.56 & 25.93 & 28.19 \\
& & MetaReg~\cite{balaji2018metareg}$^\mp$ && 42.86 & \textbf{12.68} & 32.47 & 9.37 & 43.43 & 29.87 & 28.45\\
\cline{3-11}
& & \model (\texttt{Pred-Ens})  &&50.06 & 12.23 & \textbf{34.44} & 13.07 & \textbf{46.98} & \textbf{30.13} & \textbf{31.15} \\
& & \model (\texttt{Mask-Ens})  && \textbf{50.10} & 12.17 & 34.38 & \textbf{13.14} & 46.79 & 30.01 & 31.10 \\
\midrule
& & \textbf{In-Domain} & & & & & & & & \\
\midrule
\multirow{6}{*}{\rotatebox{90}{\centering AlexNet }} & 
& Aggregate && 48.56 & 57.24 & 51.38 & 49.60 & 47.48 & 50.72 & 50.83\\
& & Aggregate-SGD$^\mp$ && 48.14 & 54.93 & 50.55 & 48.33 & 47.57 & 49.98& 49.92\\
& & Multi-Headed && 48.16 & 56.73 & 51.31 & 49.75 & 47.65 & 50.82 & 50.74 \\
& & MetaReg~\cite{balaji2018metareg}$^\mp$ && 48.87 & 56.06 & 51.23 & 49.60 & 48.66 &50.12 & 50.76\\
\cline{3-11}
& & \model (\texttt{Pred-Ens})  && 49.63 & 58.47 & 52.88 & 51.33 & 49.07 & 52.42 & 52.30 \\
& & \model (\texttt{Mask-Ens})  && 49.49 & 58.38 & 52.81 & 51.16 & 48.90 & 52.29 & 52.17 \\
& & \model-KnownDomain  && \textbf{51.91}  & \textbf{61.01}  & \textbf{54.93}  & \textbf{53.84}  & \textbf{51.08}  & \textbf{54.47}  & \textbf{54.54}  \\
\bottomrule
\end{tabular}}
\caption{\textbf{Ensembling Choices at Test-time}. We study how different
ensembling choices at test-time -- (1) \texttt{Mask-Ens}: ensemble predictions
from all the source domain masks and (2) \texttt{Pred-Ens}: combine masks
and then make a prediction -- compare in terms of in [bottom-half] an out-of-domain [top-half] performance. Using AlexNet as the backbone architecture on the DomainNet~\cite{peng2019moment} dataset, we find that
\texttt{Mask-Ens} leads to very minor ($<1\%$) drop in both in and out-of-domain performance compared to \texttt{Pred-Ens} at test-time. The columns identify the held out sixth domain for each of the multi-source shifts.$^\mp$We were unable to optimize the MetaReg~\cite{balaji2018metareg}
objective with Adam~\cite{kingma2014adam} as the optimizer and therefore,
we also include comparisons with Aggregate and MetaReg trained with SGD.}
\label{tab:ens_res_dmnt}
\end{center}
\end{table}

\begin{table}[ht!] 
\renewcommand*{\arraystretch}{1.18}
\setlength{\tabcolsep}{6pt}
\begin{center}
\resizebox{0.8\columnwidth}{!}{
\begin{tabular}{l l l  c  c c c c | c }
\toprule
& & \textbf{Method} & & A & C & P & S & Overall \\
\midrule
\multirow{15}{*}{\rotatebox{90}{\centering AlexNet }} & 
& Aggregate~\cite{li2019episodic} && 63.40 & 66.10 & 88.50 & 56.60 & 68.70 \\
& & Aggregate* && 56.20 & 70.69 & 86.29 & 60.32 & 68.38 \\
& & Multi-Headed && 61.67 & 67.88 & 82.93 & 59.38 & 67.97 \\
& & DICA~\cite{muandet2013domain} && 64.60 & 64.50 & \textbf{91.80} & 51.10 & 68.00\\
& & D-MTAE~\cite{ghifary2015domain} && 60.30 & 58.70 & 91.10 & 47.90 & 64.50 \\
& & DSN~\cite{bousmalis2016domain} && 61.10 & 66.50 & 83.30 & 58.60 & 67.40 \\
& & TF-CNN~\cite{li2017deeper} && 62.90 & 67.00 & 89.50 & 57.50 & 69.20 \\
& & Fusion~\cite{mancini2018best} && 64.10 & 66.80 & 90.20 & 60.10 & 70.30\\
& & DANN~\cite{ganin2016domain} && 63.20 & 67.50 & 88.10 & 57.00 & 69.00 \\
& & MLDG~\cite{li2018learning} && 66.20 & 66.90 & 88.00 & 59.00 & 70.00 \\
& & MetaReg~\cite{balaji2018metareg} && 63.50 &	69.50 &	87.40 & 59.10 & 69.90 \\
& & CrossGrad~\cite{volpi2018generalizing} && 61.00 & 67.20 & 87.60& 55.90 &67.90 \\
& & Epi-FCR~\cite{li2019episodic} && 64.70 & 72.30 & 86.10& 65.00 &72.00 \\
& & MASF~\cite{dou2019domain} && \textbf{70.35} & \textbf{72.46} & 90.68 & 67.33 &\textbf{75.21} \\
\cline{3-9}
& & \model (Ours)  && 64.65 & 69.88 & 87.31 & \textbf{71.42} & 73.32\\
\midrule
\multirow{11}{*}{\rotatebox{90}{\centering ResNet-18 }} & 
& Aggregate~\cite{li2019episodic} && 77.60 & 73.90 & 94.40 & 74.30 & 79.10 \\
& & Aggregate* && 72.61 & 78.46 & 93.17 & 65.20 & 77.36 \\
& & Multi-Headed && 78.76 & 72.10 & 94.31 & 71.77 & 79.24 \\
& & DANN~\cite{ganin2016domain} && 81.30 & 73.80 & 94.00 & 74.30 & 80.80 \\
& & MAML~\cite{finn2017model} && 78.30 & 76.50 & \textbf{95.10} & 72.60 & 80.60 \\
& & MLDG~\cite{li2018learning} && 79.50 & 77.30 & 94.30 & 71.50 & 80.70 \\
& & MetaReg\footnote[2]{We report the performance for MetaReg from \cite{li2019episodic} as the official PACS train/val data split changed post MetaReg publication.} \cite{balaji2018metareg} && 79.50 &	75.40 &	94.30 & 72.20 & 80.40 \\
& & CrossGrad~\cite{volpi2018generalizing} && 78.70 & 73.30 & 94.00 & 65.10 & 77.80 \\
& & Epi-FCR~\cite{li2019episodic} && \textbf{82.10} & 77.00 & 93.90 & 73.00 & \textbf{81.50} \\
& & MASF~\cite{dou2019domain} && 80.29 & 77.17 & 94.99 & 71.68 & 81.03\\
\cline{3-9}
& & \model (Ours)  && 76.90 & \textbf{80.38} & 93.35 & \textbf{75.21} & 81.46\\
\midrule
\multirow{4}{*}{\rotatebox{90}{\centering ResNet-50 }} & 
& Aggregate* && 75.49 & \textbf{80.67} & 93.05 & 64.29 & 78.38 \\
& & Multi-Headed && 75.15 & 76.37 & \textbf{95.27} & 75.26 & 80.51 \\
& & MASF~\cite{dou2019domain} && \textbf{82.89} & 80.49 & 95.01 & 72.29 & 82.67\\
\cline{3-9}
& & \model (Ours)  && 82.57 & 78.11 & 94.49 & \textbf{78.32} & \textbf{83.37}\\
\bottomrule
\end{tabular}}
\caption{\textbf{Out of Domain Generalization Results on PACS}. We compare performance (accuracy in \%) against prior work in the standard domain generalization setting of training on three domains as source and evaluating on the held-out fourth domain (identified by the column headers). We include the aggregate baseline both as reported in ~\cite{li2019episodic} as well as our own implementation (indicated as Aggregate$^*$)}
\label{tab:odres_pacs}
\end{center}
\end{table}

\subsection{More Results}
\label{sec:more_res}
In Table.~\ref{tab:odres_pacs}, we present more extensive comparisons
of \model with prior work on the PACS~\cite{li2017deeper} using AlexNet, ResNet-18 and ResNet-50 as the backbone CNN architectures. We now describe
briefly the prior approaches we compare to.

DICA~\cite{muandet2013domain} is a kernel-based optimization algorithm
that aims a learn a transformation that renders representations
invariant across domains by minimizing the dissimilarity across the
source domains. D-MTAE~\cite{ghifary2015domain} is an autoencoder based
approach which aims to learn invariant representations by cross-domain
reconstruction. DSN~\cite{bousmalis2016domain} aims to extract
representations that can be partitioned into domain-specific and 
domain-invariant components. TF-CNN~\cite{li2017deeper} learns a low-rank
parameterized CNN for end-to-end domain-generalization training. Fusion~\cite{mancini2018best} fuses predictions from all classifiers trained
on all the source domains at test-time. DANN~\cite{ganin2016domain}
leverages the source domain features extractor from Domain Adversarial
Neural Networks to generalize to target domains. MetaReg~\cite{balaji2018metareg} learns regularizers by modeling domain-shifts within the source set of distributions. MLDG~\cite{li2018learning} learns network parameters using meta-learning. Epi-FCR~\cite{li2019episodic} is a recently proposed episodic scheme to learn network parameters robust to domain-shift. MASF~\cite{dou2019domain} is a recent approach which introduces complementary losses to explicitly regularize the semantic structure of the
feature space via a model-agnostic episodic learning procedure. Cross-Grad~\cite{volpi2018generalizing} uses Bayesian Networks to
perturb the input manifold for domain generalization.

\subsection{Experimental Details}
\label{sec:exp_details}
We summarize several experimental details in this section. For all our experiments, we use Adam~\cite{kingma2014adam} as the optimizer with a batch size of 64. 
For PACS, we use an initial learning rate of $10^{-4}$ for both the network and mask
parameters decayed exponentially with a rate of 0.99 every epoch and set weight decay to
$10^{-5}$.
For DomainNet, we use an initial learning rate of $10^{-4}$ for both the network and mask
parameters decayed per-epoch using an inverse learning rate schedule\footnote{$lr_t = \frac{lr_{0}}{(1 + \gamma (t-1))^p}$ where $\gamma=10^{-4}, p=0.75$, $t$ identifies the epoch and $lr_0$ is the initial learning rate.}
and set weight decay to 0.
We conduct a sweep over values of
$\lambda_O$ -- coefficient of the \texttt{sIoU} loss -- in the range
$\{0, 10^{-5}, 10^{-4}, 10^{-3}, 10^{-2}, 10^{-1}, 1\}$. Our backbone CNN architectures are initialized
with ImageNet~\cite{krizhevsky2012imagenet} pretrained checkpoints.
We initialize the final linear layer weights (to be learned from scratch) from a zero centered normal distribution
($\mathcal{N}(0, 0.001)$) and a uniform distribution (standard in PyTorch) for DomainNet and PACS respectively.
For
all our experiments, we initialize the mask parameters from the uniform
distribution, i.e., $\Tilde{\mathbf{m}}^d \sim \mathtt{U}(0,1)$. 
We select the best checkpoints across 50 epochs of training based on overall in-domain validation
accuracy.
We implement
everything in the Pytorch~\cite{NEURIPS2019_9015} framework\footnote{The authors of~\cite{peng2019moment} indicated in communication that they used Caffe to implement
the multi-source baselines. We re-implement the multi-source baselines in
PyTorch~\cite{NEURIPS2019_9015} to ensure consistency across all our reported results. The subsequent differences in multi-source baseline accuracies can be attributed to the differences in how 
AlexNet 
is implemented in PyTorch and Caffe.}. Our code is available at \href{https://github.com/prithv1/DMG}{https://github.com/prithv1/DMG}


\end{document}